\documentclass[authoryear,12pt]{elsarticle}
\usepackage{lipsum}
\makeatletter
\def\ps@pprintTitle{%
 \let\@oddhead\@empty
 \let\@evenhead\@empty
 \def\@oddfoot{}%
 \let\@evenfoot\@oddfoot}
\makeatother

\usepackage[T1]{fontenc}
\usepackage[utf8]{inputenc}
\usepackage{amssymb}
\usepackage{adjustbox}
\usepackage{amsmath, amsthm, bbm}
\usepackage{hyperref}
\usepackage{float}
\usepackage{comment}
\usepackage{graphicx}
\usepackage[usenames,dvipsnames]{xcolor}
\usepackage{algorithm, algpseudocode}
\usepackage{hyperref}
\usepackage{multirow}
\usepackage{booktabs}
\usepackage{mathtools}

\usepackage{tikz}

\newcommand{\x}{\mathbf{x}}
\newcommand{\X}{\mathbf{X}}

\renewcommand{\P}{\mathbb{P}}
\newcommand{\R}{\mathbb{R}}

\newcommand{\E}{\mathbb{E}}
\newcommand{\V}{\mathbb{V}}

\newcommand{\mT}{\mathcal{T}}

\newtheorem{Theorem}{Theorem}
\newtheorem{Assumption}{Assumption}
\newtheorem{Lemma}{Lemma}
\newtheorem{Corollary}{Corollary}
\newtheorem{Definition}{Definition}

\newtheorem{Example}{Example}
\usepackage{enumitem}

\def\sA{{\mathcal{A}}}

\def\sX{{\mathcal{X}}}
\def\sY{{\mathcal{Y}}}

\newcommand{\RNum}[1]{\uppercase\expandafter{\romannumeral #1\relax}}
\usepackage[textfont=bf]{caption}  
\usepackage{subcaption}
\usepackage{tikz}
\usepackage{color}
\usepackage{soul}

\renewcommand{\t}{\textup{\textrm{t}}}
\newcommand{\train}{\textrm{train}}
\renewcommand{\cal}{\textrm{cal}}
\renewcommand{\part}{\textrm{part}}
\newcommand{\cut}{\textrm{cut}}
\renewcommand{\split}{\textrm{split}}
\newcommand{\local}{\textrm{local}}

\newcommand{\locart}{\textup{\texttt{Locart}}}
\newcommand{\alocart}{\textup{\texttt{A-Locart}}}
\newcommand{\loforest}{\textup{\texttt{Loforest}}}
\newcommand{\aloforest}{\textup{\texttt{A-Loforest}}}
\newcommand{\wloforest}{\textup{\texttt{W-Loforest}}}
\newcommand{\id}[1]{\mathbbm{1}\{ #1 \}}

\journal{Information Sciences}

\begin{document}

\begin{frontmatter}



\title{Regression Trees for Fast and Adaptive Prediction Intervals}

\affiliation[inst1]{organization={Department of Statistics, Federal University of S\~{a}o Carlos},
            city = {S\~{a}o Carlos},
            postcode = {13566-590}, 
            state = {S\~{a}o Paulo},
            country = {Brazil}}

\affiliation[inst2]{organization={Institute of Mathematical Sciences and Computation, University of S\~{a}o Paulo},
            city = {S\~{a}o Carlos},
            postcode = {13565-905}, 
            state = {S\~{a}o Paulo},
            country = {Brazil}}

\affiliation[inst3]{organization = {Institute of Mathematics and Statistics, University of S\~{a}o Paulo},
city = {S\~{a}o Paulo},
postcode = {05508-090},
state = {S\~{a}o Paulo},
country = {Brazil}}

\author[inst1,inst2]{Luben M. C. Cabezas}
\author[inst1,inst2]{Mateus P. Otto}
\author[inst1]{Rafael Izbicki}
\author[inst3]{Rafael B. Stern}

\begin{abstract}
Predictive models make mistakes. Hence, there is a need to quantify the uncertainty associated with their predictions. Conformal inference has emerged as a powerful tool to create statistically valid prediction regions around point predictions, but its naive application to regression problems yields non-adaptive regions. New conformal scores, often relying upon quantile regressors or conditional density estimators, aim to address this limitation. Although they are useful for creating prediction bands, these scores are detached from the original goal of quantifying the uncertainty around an \emph{arbitrary} predictive model. This paper presents a new, model-agnostic family of methods to calibrate prediction intervals for regression problems with local coverage guarantees. Our approach is based on pursuing the coarsest partition of the feature space that approximates conditional coverage. We create this partition by training regression trees and Random Forests on conformity scores. Our proposal is versatile,  as it applies to various conformity scores and prediction settings and demonstrates superior scalability and performance compared to established baselines in simulated and real-world datasets. We provide a Python package \href{https://github.com/Monoxido45/clover}{\texttt{clover}} that implements our methods using the standard \textit{scikit-learn} interface.
\end{abstract}


\end{frontmatter}


\section{Introduction}
\label{sec:intro}

A large body of literature on regression analysis has been devoted to developing models to create point prediction for a label $Y$ based on features $\x$. However, in many applications, point predictions are not enough: practitioners also need to have a measure of uncertainty around the predicted value $\hat{\mu}(\x)$. Several methods to create prediction sets around $\hat{\mu}(\x)$ have been developed, but these sets traditionally
require strong assumptions about the data-generating process to have the correct coverage
\citep{butler1980predictive,neter1996applied,steinberger2016leave}. 

 More recently, conformal methods \citep{papadopoulos2002inductive,Vovk2005,Vovk2009,Lei2018} have emerged as a powerful tool to create prediction sets. 
 A noteworthy characteristic of conformal methods is their ability to control the marginal coverage of prediction bands in a distribution-free setting. In other words, assuming only that the instances $(\X,Y)$ are independently and identically distributed (i.i.d.), conformal prediction regions $C(\X)$ possess the property that, for a fixed $\alpha \in (0,1)$ set by the user, 
 \begin{equation*}
   \P[Y \in C(\X)] = 1-\alpha
 \end{equation*}
even when the model used to construct $C$ is misspecified. 

However, achieving marginal coverage alone is often insufficient. 
Ideally, prediction regions should also be local. This is not always the case for conformal methods. For instance,
the \emph{conditional coverage} of conformal regions, given by  $\P[Y \in C(\X) | \X=\x]$, can deviate significantly from $1-\alpha$ at various locations $\x$. This is undesirable in practice as it implies that prediction sets may be well-calibrated overall but fail to cover the true $Y$ values for certain subgroups \citep{barocas2017fairness}.  Such lack of calibration within specific groups raises serious fairness concerns, potentially leading to biased decision-making
\citep{kleinberg2016inherent,zhao2020individual}.

Unfortunately, in the distribution-free setting, it is impossible to construct predictive intervals $C(\X)$ based on a finite number of samples that attain conditional coverage and have reasonable lengths \citep{Lei2014, barber2021limits}. Some conformal methods bypass this negative result in the asymptotic regime and achieve conditional coverage \emph{asymptotically} \citep{romano2019conformalized,izbicki2022cd}. Most of them, however, require new conformity scores that have no relationship to $\hat{\mu}(\x)$ (see Section~\ref{sec:relwork}), and thus cannot be used to construct prediction intervals around the estimated regression function. 

Thus, a key question is how to effectively construct local prediction intervals based on the estimated regression function.

\subsection{Novelty and Contributions}

In this paper, we propose a hierarchy of methods to construct local prediction intervals around point estimates of a regression function  $\hat \mu(\x)$. That is, 
our confidence sets  have  the shape
\begin{equation*}
  C(\x) = (\hat{\mu}(\x) - \hat{t}_{1-\alpha}(\x), \hat{\mu}(\x) + \hat{t}_{1-\alpha}(\x)).
\end{equation*}

Our methods interpolate between conformal and non-conformal procedures, yield calibrated intervals with theoretical guarantees, and exhibit improved scalability compared to literature baselines.

Our first method, which serves as a starting point in the hierarchy, is \locart{}. 
\locart{} is based on creating  a partition $\sA$ of the feature space,
and defining the cutoffs $\hat{t}_{1-\alpha}(\x)$ 
by separately applying conformal prediction to each element of $\sA$.
 Contrary to conventional methodologies (refer to Section \ref{sec:relwork} for an overview of related works), the selection of $\sA$ is guided by a data-driven optimization process designed to yield the most accurate estimates of the cutoffs.
 Specifically,
 the partition $\sA$ is such that the output set of \locart{}, $C_{\locart{}}(\X)$, satisfies
$$\P[Y \in C_{\locart{}}(\X)|\X \in A] \approx \P[Y \in C_{\locart{}}(\X)|\X=\x]$$ 
and guarantees local coverage,
 \begin{equation*}
    \P[Y \in C_{\locart{}}(\X)| X \in A] \ge 1-\alpha,
\end{equation*}
for every $A \in \sA$.
In other words, we expect \locart{} to have good conditional coverage.


In the next hierarchy level, we have \loforest{}. \loforest{} builds on \locart{} by using multiple regression trees on conformity scores to define its prediction interval. That is, \loforest{} is a Random Forest of trees on conformal scores. Although \loforest{} is not a conformal method, it shows excellent conditional coverage in practice (see Section~\ref{sec:experiments}).


\subsection{Relation to other work}
\label{sec:relwork}

There exist many non-conformal approaches for uncertainty quantification in regression. These include Gaussian Processes \citep{rasmussen2005gaussian}, quantile regression \citep{koenker1978regression, koenker2005quantile, meinshausen2006quantile}, the bootstrap, the jackknife, cross-validation \citep{efron1983leisurely}, out-of-bag uncertainty estimates \citep{breiman1996out}, Bayesian neural networks \citep{neal2012bayesian}, dropout-based approximate Bayesian inference \citep{gal2016dropout}, and others (see \citeauthor{dey2022calibrated}, \citeyear{dey2022calibrated}; \citeauthor{dewolf2022valid}, \citeyear{dewolf2022valid}). In general, however, these methods fail to provide even marginally valid predictive intervals in the finite-sample setting (see, for instance, \cite{barber2021predictive} for the failure of jackknife and cross-validation). 

In turn, a primary feature of all split conformal inference methods is that they always yield marginally valid prediction intervals \citep{papadopoulos2008normalized, Vovk2012, Lei2014,valle2023quantifying}. Moreover, beyond marginal coverage, several conformal methods construct conditionally valid prediction intervals in the asymptotic regime or locally valid prediction intervals in the finite-sample setting. Methods falling into the first category often leverage consistent quantile regression estimators or rely on the design of sophisticated conformal scores whose conditional distribution (on $\X$) is independent of $\X$. This is the case, for instance, of conformalized quantile regression \citep{romano2019conformalized}, distributional conformal prediction \citep{chernozhukov2019distributional}, Dist-split \citep{izbicki2020}, and HPD-split \citep{izbicki2022cd}. Nevertheless, different than ours, these methods are not based on estimates of the regression function $\hat{\mu}(\x)$, which makes them unsuitable for building prediction intervals centered at $\hat{\mu}(\x)$. 

\begin{figure}[!htb]
    \centering
    \includegraphics[width=\textwidth]{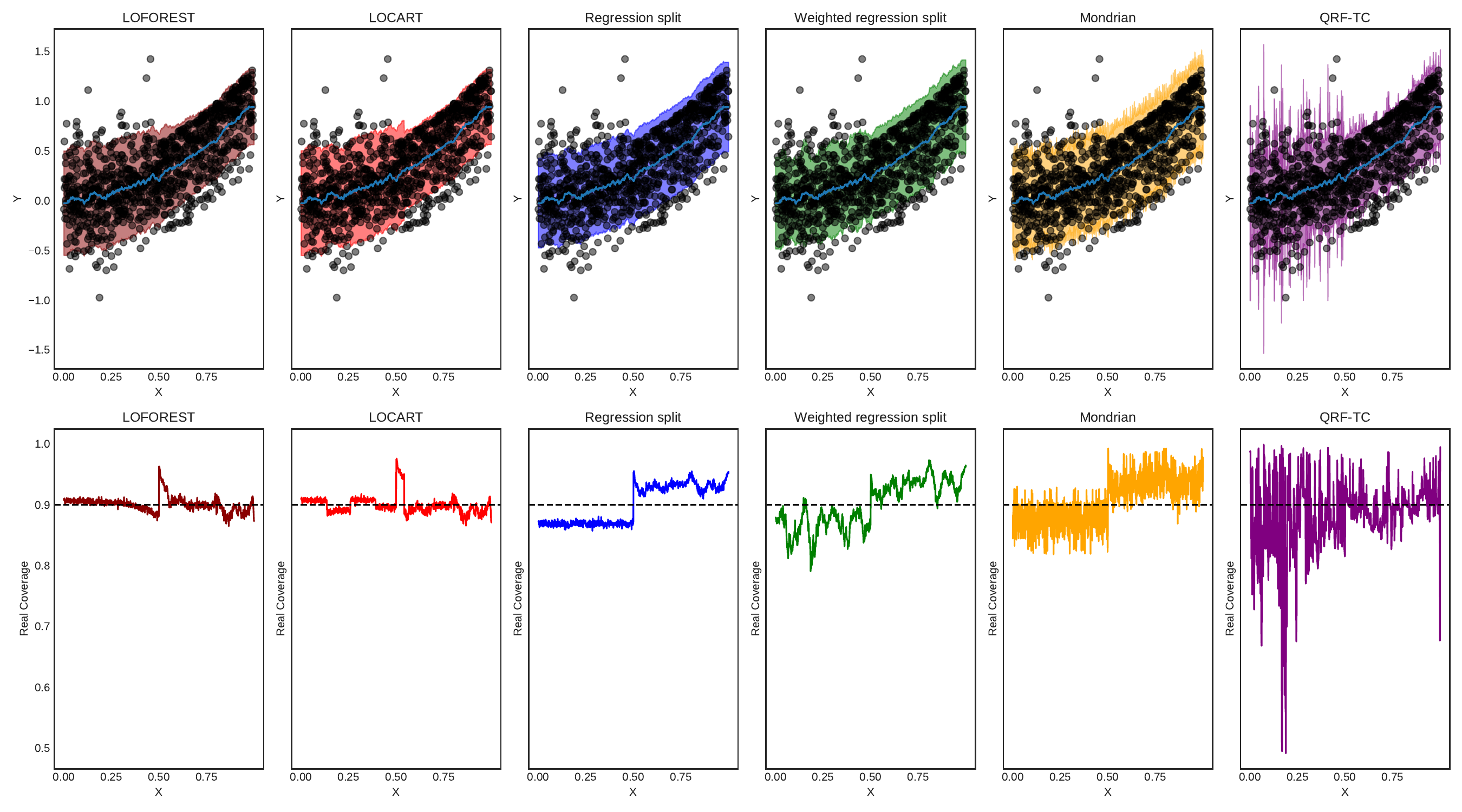}
    \includegraphics[width=\textwidth]{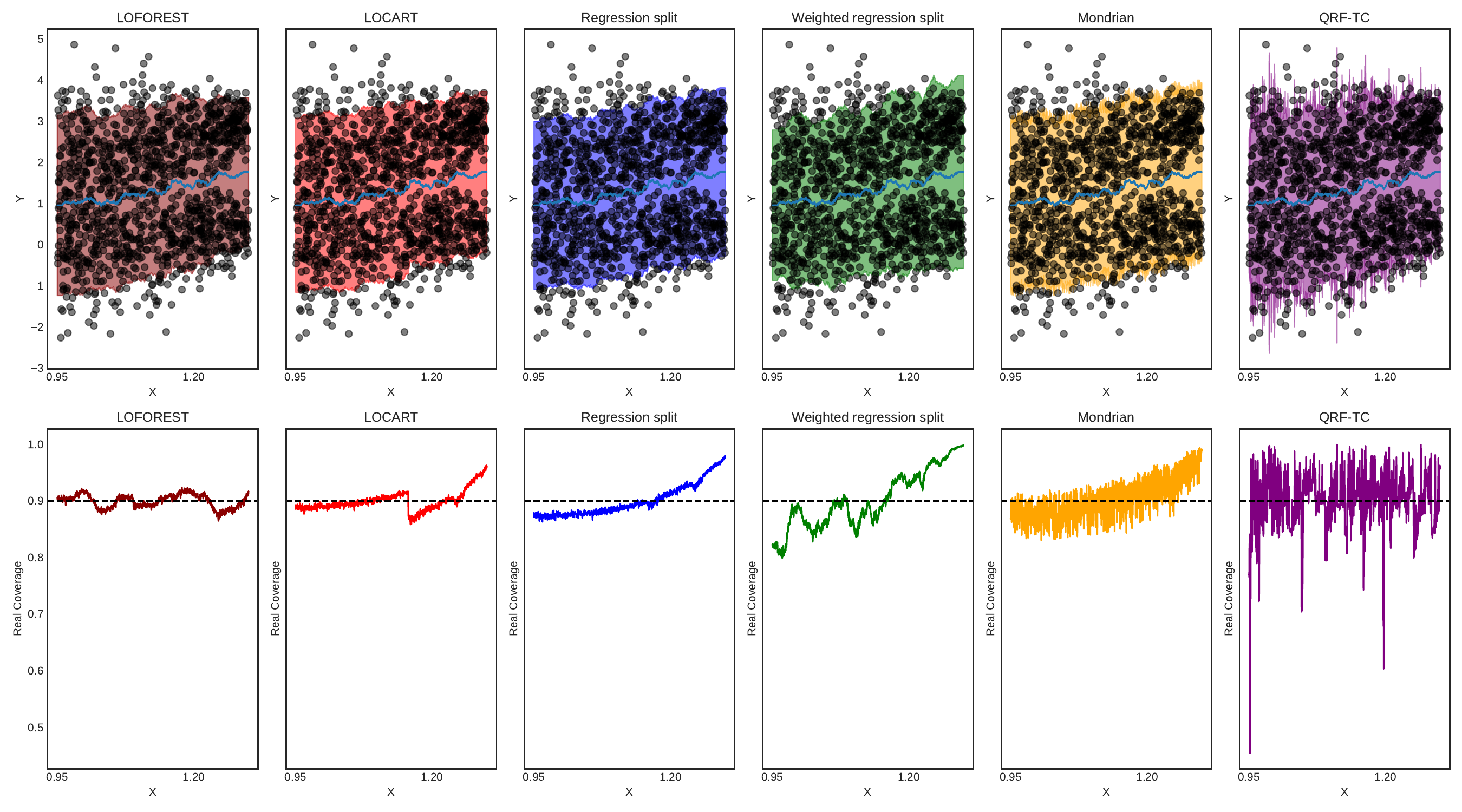}
    \caption{Top: failure of locally weighted regression split: the conditional mean absolute deviation of $[Y-\hat{\mu}(X)]$ is constant, but the conditional quantiles of $|Y-\hat{\mu}(X)|$ are not. Bottom: failure of using ``difficulty'' as the conditional variance of $Y$: the conditional variance $\V(Y|X=x)$ is constant, but the conditional quantiles of $|Y-\hat{\mu}(X)|$ are not. In both cases, the intervals do not adapt to the data's local features and would fail to guarantee asymptotic conditional coverage.
    Our methods (\locart{} and \loforest), however, have good performance.}
    \label{fig:adversary}
\end{figure}

A common goal of the aforementioned methods is to make prediction intervals adapt to the local structure of the data. In the regression setting, this is often sought by normalizing conformal scores with point-wise model uncertainty estimates. This procedure, known as normalization \citep{papadopoulos2008normalized, papadopoulos2011regression, johansson2014regressiona} or locally weighted conformal prediction \citep{Lei2018}, is similar to ours only in the sense we also accomplish data-adaptivity. Our approach, however, seeks adaptation with an optimal level of ``granularity'', which, as discussed in Section~\ref{sec:methodology}, is not on the level of instances or point-wise. We give an example in Figure~\ref{fig:adversary} where the locally weighted regression split method \citep{Lei2018} with a perfectly estimated ``normalizer'' fails to provide prediction intervals that are asymptotically conditionally valid. See ~\ref{appendix:baselines} for details.

In fact, seeking adaptivity of prediction intervals on the level of instances is unfeasible, as constructing non-trivial conditionally valid prediction intervals is inherently hard in the finite-sample setting  \citep{Lei2014, barber2021limits}. As a consequence, many approaches have instead focused on defining sensible collections of subsets (e.g., partitions) of the feature space. This is the case of \cite{Vovk2012}, \cite{Lei2014},  \cite{bostrom2020mondrian},  \cite{bostrom2021mondrian}, and \cite{barber2021limits}. For instance, \citeauthor{Lei2014} partitions the feature space by creating a hyper-rectangular mesh, while Mondrian-based approaches \citep{bostrom2020mondrian, bostrom2021mondrian} define a partition with a predefined Mondrian taxonomy \citep{Vovk2005}. For both these methods, the split conformal approach is applied to each partition element to control local coverage similarly to ours. Notwithstanding, our approach is markedly different since the partition is derived in a data-adaptive manner based on the natural partition induced by training a regression tree of conformity scores. We remark that, although \cite{papadopoulos2010neural, Lei2018} train regression models for the conformal scores, this is done within the normalization framework and thus inherits its limitations. Meanwhile, Mondrian taxonomies induced by binning certain uncertainty estimates \citep{bostrom2020mondrian}, such as the conditional variance of the response \citep{bostrom2017accelerating}, are suboptimal and fail to produce prediction intervals with asymptotic conditional coverage (see the example in Figure~\ref{fig:adversary} and details on~\ref{appendix:baselines}). Finally, we are motivated by a goal similar to \cite{ding2023class}, where the authors sought to cluster instances with a similar distribution of conformal scores. Nonetheless, their clustering scheme is class-conditional (i.e., $Y$-conditional), while ours is strictly $\X$-conditional. 

Another framework for conformal inference in regression is localized conformal prediction (LCP) \citep{guan2021localized}. LCP is based on weighting calibration samples according to their similarity to the test sample, as measured by a localizer. These similarity measures are then used to compute an estimate of the conditional (on $\X$) distribution of scores. In practice, because it requires storing the similarity matrix and recomputing the miscoverage level to rectify the induced non-exchangeability, LCP falls short in computational scalability. Within the LCP framework, LCP-RF \citep{amoukou2023adaptive} is especially kindred to our approach. LCP-RF leverages the adaptive nearest neighbor \citep{lin2006random} characterization of Quantile Regression Forests \citep{meinshausen2006quantile} to estimate the conditional distribution of scores. Two significant distinctions set us apart. First, we do not include the test sample in the estimate of the conditional distribution of scores, and thus, we do not have to adjust the miscoverage level (a trademark of LCP-based methods). Second, we do not construct the prediction intervals based on the estimated quantile of the Quantile Regression Forest. Instead, we estimate quantiles for every tree with a Quantile Regression Forest fitted on that tree and average these quantiles over all trees in the forest. 

\subsection{Notation and Organization}

For an integer $n \ge 1$, we write $[n]$ to denote the set $\{1, \dots, n\}$. We will denote by $\sqcup$ the union of disjoint sets. We use $\mathbbm{1}\{ \cdot \}$ for the indicator function. In addition, we denote by $\hat{q}_{\phi}(A)$ the empirical $\phi$-quantile of the set $A = \{a_1, \dots, a_n\}$; that is,
\begin{equation*}
    \hat{q}_{\phi}(A)= \inf \left\{ t : \frac{1}{n} \sum_{i=1}^n \mathbbm{1}\{ a_i \le t\} \ge \phi \right\}.
\end{equation*}

The remainder of the paper is organized as follows. We give an overview of Conformal Inference and Regression Trees/Forests in Section~\ref{sec:preliminaries}. In Section~\ref{sec:methodology} we introduce our methods, discussing their properties and theoretical guarantees. In Section~\ref{sec:experiments}, we validate our approach in simulated and real-world datasets, comparing it to baselines. We conclude with a brief discussion in Section~\ref{sec:final}. All proofs are deferred to \ref{appendix:proofs}. 

\section{Preliminaries}
\label{sec:preliminaries}


\subsection{Conformal Inference}
\label{subsec:conformal_pred}
The main ingredient of the conformal inference framework is the conformity score $\hat{s}: \sX \times \sY \to \R$, which measures how well the outputs of a regression or classification model accurately predict new labels $Y \in \sY$. In the regression setting, given an estimator $\hat{\mu}(\cdot)$ of the regression function $\E[Y|\X=\x]$, a standard choice is the regression residual $\hat{s}(\x, y) = |y - \hat{\mu}(\x)|$. 
Similarly, for the classification setting, given an estimator $\hat{\pi}_Y(\x)$ of the true class probabilities $\pi_Y(\x)$, one can use the generalized inverse quantiles from adaptive prediction sets \citep{romano2020classification} or its regularized version \citep{angelopoulos2020uncertainty}. In what follows, to avoid unnecessary restriction to either regression or classification, we let the underlying predictive model implicit in the choice of the conformity score. In this sense, ``training'' a conformal score refers to training this underlying predictive model and plugging it into the definition of the chosen conformity score.\footnote{We write $\hat{s}$ (with hat) to reinforce this dependence on an underlying trained model.}

A general method for obtaining prediction intervals in conformal inference with distribution-free guarantees is the split conformal prediction (split CP) framework \citep{ papadopoulos2002inductive, Vovk2005}.
Considering $\{ (\X_i, Y_i) \}_{i=1}^{n+1}$ to be exchangeable data, split CP works as follows:

\begin{enumerate}
    \item Split the data indices into two non-empty disjoint sets $I_{\train}$ and $I_{\cal}$, such that $I_{\train} \sqcup I_{\cal} = [n]$.  

    \item Train a conformity score $\hat{s}: \sX \times \sY \to \R$ on  $ \{ (\X_i, Y_i) \}_{i \in I_{\train}}$.

    \item Evaluate $\hat{s}(\X_i, Y_i)$ for all $i \in I_{\cal}$.

    \item Calculate $s^*$ as the empirical $\left(1+ 1/n_{\cal} \right)(1-\alpha)$-quantile of $\hat{s}_i = \hat{s}(\X_i, Y_i)$ for $i \in I_{\cal}$, where $n_{\cal}$ is the size of $I_{\cal}$.

    \item For the new instance $(\X_{n+1}, Y_{n+1})$, define the split conformal prediction set as
    \begin{equation}
        C_{\split}(\X_{n+1}) = \{ y \in \sY : \hat{s}(\X_{n+1}, y) \le s^*\}.
        \label{eq:splitinterval}
    \end{equation}
\end{enumerate}

Because of the exchangeability of the data, the splitting step guarantees that the scores $\hat{s}_i$ for $i \in I_{\cal} \sqcup \{n+1\}$ are also exchangeable. As a consequence, the rank of $\hat{s}(\X_{n+1}, Y_{n+1})$ among these $n+1$ conformity scores is drawn uniformly from the $n+1$ possible indices. By choosing a cutoff $s^*$ as the empirical $\left(1+ 1/n_{\cal} \right)(1-\alpha)$-quantile of the scores, we get that $\P[Y_{n+1} \in C_{\split}(\X_{n+1})]  = \P[\hat{s}(\X_{n+1}, Y_{n+1}) \le s^*] \ge 1-\alpha$. The following theorem formalizes this explanation. Furthermore, it asserts that conditional on the training data (used to construct $\hat{s}$) and on average over all possible calibration datasets with $n_{\cal}$ elements, the prediction band $C_{\split}$ is not too wide provided $n_{\cal}$ is sufficiently large. (See \citep{Vovk2012, bian2023training} for guarantees incorporating the randomness of both the training and calibration datasets).

\begin{Theorem}[\citeauthor{Lei2018}, \citeyear{Lei2018}] Given an exchangeable sequence $(\X_i, Y_i)_{i=1}^{n+1}$ and a miscoverage level $\alpha \in (0, 1 )$,
\begin{equation*}
    \P[Y_{n+1} \in C_{\split}(\X_{n+1})] \ge 1-\alpha,
\end{equation*}
where $C_\textrm{split}$ is the prediction band from the split conformal framework~\eqref{eq:splitinterval}. In addition, if the conformal scores are almost surely distinct, then
\begin{equation*}
    \P[Y_{n+1} \in C_{\split}(\X_{n+1})] \le 1-\alpha + \frac{1}{n_{\cal}+1}.
\end{equation*}
\end{Theorem}

The split conformal framework can be naturally extended to produce prediction bands with local validity \citep{Vovk2005, Lei2014}. Let $\sA = \{ A_1, \dots, A_K \}$ be a finite partition of the feature space and consider $\mT: \sX \to \sA$ the function that maps $\X$ to the partition element it belongs. The extension works as follows: given the initial split $I_{\train}$ and $I_{\cal}$, we further split $I_{\cal}$ into the sets $I_j = \{ i \in I_\cal: \mT(\X_i) = A_j \}$ for $j=1, \dots, K$.  As before, the splitting step guarantees that the scores $\hat{s}_i$ for $i \in I_j \sqcup \{n+1\}$ are exchangeable. In complete analogy to split CP, for each split $I_j$, we calculate $s^*_j$ as the empirical $(1+1/n_j)(1-\alpha)$-quantile of the scores $\hat{s}_i$ for $i \in I_j$, where $n_j$ is the size of $I_j$. Thus, we obtain that $\P[Y_{n+1} \in C_{\local}(\X_{n+1}) | \X_{n+1} \in A_j]  \ge 1-\alpha$, where
\begin{equation}
    C_{\local}(\X_{n+1}) = \{ y \in \sY: \hat{s}(\X_{n+1}, y) \le s^*_{j} \}
    \label{eq:splitinterval-local}
\end{equation}
is the local split conformal prediction set. If $I_j$ is an empty set, we take the convention that $s^* = + \infty$. Theorem~\ref{theorem:local-validity-splitcp} formalizes this discussion by stating the local validity of the prediction intervals \eqref{eq:splitinterval-local}.

\begin{Theorem}
\label{theorem:local-validity-splitcp}
Let $\sA = \{A_1, \dots, A_K\}$ be a finite partition of the feature space. Given an exchangeable sequence $(\X_i, Y_i)_{i=1}^{n+1}$ and a miscoverage level $\alpha \in (0, 1 )$,
\begin{equation*}
    \P[Y_{n+1} \in C_{\local}(\X_{n+1}) | \X_{n+1} \in A_j] \ge 1-\alpha,
\end{equation*}
for all $j = 1, \dots, K$, where $C_{\local}$ is the prediction band in~\eqref{eq:splitinterval-local}. 
\end{Theorem}

Applying the split CP framework to each partition element entails a cost. If partition elements' are not well-populated, prediction intervals grow wide. To see this, suppose we are given a realization of the dataset. If $\alpha = 0.05$, then any partition element containing less than $\bar{n} = 19$ instances will have trivial local bands, with $C_{\local}(\cdot) = \R$. Therefore, the extension of split CP that attains local validity must be combined with a partition of the feature space whose elements are guaranteed to contain sufficiently many instances of the calibration split.

\subsection{Regression Trees}
\label{sec:regression_trees}

Suppose we want to create an estimator $\hat{\mu}(\x)$ of the regression function $\E[Y|\X= \x]$ given access to i.i.d samples $\{ (\X_i, Y_i) \}_{i=1}^n$. Regression trees builds $\hat{\mu}$ by partitioning the feature space $\sX$ in sets of the form $\{ X_j \le \ell\}$ and $\{ X_j > \ell\}$, where $X_j$ is a feature and $\ell$ is a level. The splitting features and levels are chosen to minimize the sample variance of $Y$ inside each partition element. However, a ``global'' partitioning strategy is unfeasible due to the combinatorial hardness of testing all possible combinations of splitting features and levels. Instead, the partition is chosen by optimizing a greedy or local objective. 

A standard way to greedily grow the tree is through the CART methodology \citep{breiman1993classification}. Given a  node $\t$, starting from $\t = \t_1 = \sX$ in the first step, the CART algorithm operates as follows:

\begin{enumerate}
    \item Consider the left and right splits of $\t$,
    \begin{equation*}
        \t_L(j, \ell) = \{ \X  \in \t: X_j \le \ell \} \quad\textrm{and}\quad \t_R(s, \ell) = \{ \X  \in \t: X_j > \ell\}.
    \end{equation*}

    \item Determine the split $(\hat{\jmath}, \hat{\ell})$ such that 
    \begin{equation*}
      \frac{1}{N(\t)}\Biggl( \sum\limits_{\X_i \in \t_L} (Y_i - \overline{Y}_{\t_L})^2 + \sum_{\X_i \in \t_R} (Y_i - \overline{Y}_{\t_R})^2\Biggr)
    \end{equation*}
    is minimized, where $N(\t) = | \{ \X \in \t\} |$ and $\overline{Y}_\t = \frac{1}{N(\t)} \sum_{\X_i \in \t} Y_i$.

    \item Recurse the procedure on $\t_L$ and $\t_R$ until the maximum depth $K$ is reached or some other stopping criterion is satisfied.
\end{enumerate}

If the maximum depth $K$ of the tree is set to a high value, trees may overfit. This occurs after successive splittings of the training data since each terminal node of a tree may contain a single data instance. To avoid this, the tree may be pruned as the training unrolls or afterward. In the former case, called pre-pruning, the growth of the tree is controlled by setting the minimal amount of samples in each of its internal or terminal nodes (via the \texttt{min\_samples\_split} and \texttt{min\_samples\_leaf} hyperparameters, respectively, in the \textit{scikit-learn} \citep{pedregosa2011scikit} implementation). 
In the latter case, called post-pruning, subtrees or nodes are removed until a balance between model complexity and predictive performance is reached. 

After training and pruning, the tree has a set of terminal nodes (or leaves) $\tilde{T}$ that can be naturally identified as a partition of the feature space. With this, define $\mT: \sX \to \tilde{T}$ as the function that maps a new instance $\x_{n+1} \in \sX$ to the terminal node (partition element) it belongs. Then, the regression function estimate at $\x_{n+1}$ is $\hat{\mu}(\x_{n+1}) = \overline{Y}_{\mT(\x_{n+1})}$. It is also possible to view this estimate as a weighted nearest-neighbor method \citep{lin2006random}; defining the weighting function
\begin{equation*} 
    w_i(\x) = \frac{\id{ \X_i \in \mT(\x)}}{N(\mT(\x))},
\end{equation*}
the regression function estimate can be rewritten as
\begin{equation*}
    \hat{\mu}(\x_{n+1}) = \sum_{i=1}^n w_i(\x_{n+1}) Y_i,
\end{equation*}
which highlights the data-adaptive nature of regression trees since the weight function is higher for the training points closer to the new instance $\x_{n+1}$. This property allows one to construct tree-based estimators of quantities beyond the conditional mean \citep{cevid2022distributional}. 
Indeed, \citet{meinshausen2006quantile} shows that, under some mild assumptions, the function
\begin{equation}
\label{eq:qrf}
    \hat{F}(y|\X=\x_{n+1}) = \sum_{i=1}^n w_i(\x_{n+1}) \id{Y_i \le y}
\end{equation}
is a consistent estimator of the conditional distribution $Y|\X=\x$ for all $\x \in \sX$. Although the rate of convergence of $\hat{F}$ to the true distribution function might be possibly improved by bagging multiple trees, as done in Quantile Regression Forests \citep{meinshausen2006quantile}, the consistency holds even for a single tree. We will explore this fact to show the asymptotic conditional coverage property of \locart{}.

\section{Methodology}
\label{sec:methodology}

In this section, we introduce our procedures for calibrating predictive intervals. First, building on the construction of the oracle prediction interval, we present an optimal partitioning scheme for local conformal. Then, we propose Local Coverage Regression Trees (\locart{}) as a realization of this scheme. Next, we introduce Local Coverage Regression Forests (\loforest{}), which is a non-conformal alternative to \locart{} based on averaging cutoff estimates that exhibit improved empirical performance.  Then, we introduce the augmented versions of both algorithms, \alocart{} and \aloforest{}, that can further refine the partitions by augmenting the feature space. Finally, we present a weighted version of \loforest{} called \wloforest{} that can be used to improve the locally weighted prediction intervals \citep{Lei2018}.

\subsection{Motivation} 
\label{sec:motivation}

We consider prediction sets of the form
\begin{equation*}
  C(\x) = (\hat{\mu}(\x) - \hat{t}_{1-\alpha}(\x), \hat{\mu}(\x) + \hat{t}_{1-\alpha}(\x)).
\end{equation*}
Ideally, we would like to choose the cutoff $\hat{t}_{1-\alpha}$ such that, for a fixed miscoverage level $\alpha \in (0,1)$, the prediction set $C(\x)$ controls conditional coverage.
The prediction set that uses this optimal cutoff  is named the \emph{oracle prediction interval}:
\begin{Definition}[Oracle prediction interval]
\label{def:oracle_PI} Given a regression estimator $\hat{\mu}$, the oracle prediction interval $C^*(\x)$ is defined as
    \begin{align*}
        C^*(\x) := (\hat{\mu}(\x) - t_{1-\alpha}^*(\x), \hat{\mu}(\x) + t_{1-\alpha}^*(\x)), 
    \end{align*}
    where $t_{1-\alpha}^*(\x)$ is the $(1-\alpha)$-quantile of the regression residual $\hat{s}(\X, Y) = |Y-\hat{\mu}(\X)|$ conditional on $\X = \x$. This is the only symmetric interval around $\hat{\mu}(\cdot)$ such that
    \begin{align*}
        \P[Y \in C^*(\X)|\X = \x,\hat{\mu}] = 1 - \alpha.
    \end{align*}    
\end{Definition}

It is of course impossible to obtain $t^*_{1-\alpha}$
in practice. Our goal is, therefore, to get as close as possible to $t^*_{1-\alpha}$.
We will do this by 
calibrating the cutoff $\hat{t}_{1-\alpha}(\x)$
using a local conformal approach.

To recover $t^*_{1-\alpha}$ effectively, the partition $\sA$ must be designed such that all $\x$ values within the same element $A \in \sA$ (approximately) share the same optimal cutoff. This is because a local conformal method, by design, will yield the same cutoff for all $\x$ values belonging to the same element $A \in \sA$. In other words, if $A$ is such that all $\x \in A$ approximately share the same $t^*_{1-\alpha}$, then
 $$\P[Y \in C(\X)|\X=\x,\hat{\mu}] \approx \P[Y \in C(\X)|\X \in A,\hat{\mu}] =1-\alpha,$$
that is, the local conformal intervals $C(\X)$ will not only have local coverage but will also be close to conditional coverage.

Euclidean partitions are one approach that aims to achieve this objective \citep{Lei2014}: as $\text{dim}(A) \longrightarrow 0$, $A$ will only have a single point $\x$, and therefore all elements of $A$ will share the same optimal cutoff. 
Euclidean partitions, however, are not optimal: two $\x$'s may be far from each other but still share the same optimal cutoff, as the next examples show:

\begin{Example}\textbf{[Location family]}
 Let  $h(y)$ be a density, 
 $\mu(\x)$ a function, and
 $Y|\x \sim h(y-\mu(\x))$.
 In this case,
 $t^*_{1-\alpha}(\x_a)=t^*_{1-\alpha}(\x_b)$,
 for every $\x_a,\x_b \in \mathbb{R}^d$.
 For instance, if $Y|\x\sim N(\beta^t \x,\sigma^2)$, 
 then all instances have the same optimal cutoff.
 Thus, in this special scenario,
  a unitary partition $\sA=\mathcal{X}$ would already lead to 
 conditional validity as the calibration sample increases.
\end{Example}

\begin{Example}\textbf{[Irrelevant features]}
 If $\x_S$ is a subset of the features such that the conditional densities satisfy
 $f(y|\x)=f(y|\x_S)$, then
 $t^*_{1-\alpha}(\x)=t^*_{1-\alpha}(\x_S)$, that is, the optimal cutoff does not depend
 on the irrelevant features, $S^c$.
  These irrelevant features, however, 
 affect the Euclidean distance between $\x$'s.
\end{Example}

Therefore, even if two $\x$ values are distant, they can still be included in the same $A$ if they share the same $t^*_{1-\alpha}(\x)$.
 This is equivalent to saying that the optimal partition should group instances with the same conditional distribution of the residuals $\hat s(\X,Y)| \X = \x$.
 

Theorem \ref{thm:optimal_cutoff_eq_relation} formalizes the construction and optimality of such partition.
\begin{Theorem}[The coarsest partition with same oracle cutoffs]
\label{thm:optimal_cutoff_eq_relation}
    Let $\mathcal{A}$ be a partition of $\sX$ such that, for any $A \in \mathcal{A}$, $\x_1, \x_2 \in A$ if and only if $\hat{s}(\X, Y) | \X = \x_1 \sim \hat{s}(\X, Y) | \X = \x_2$, where $\hat{s}(\X, Y) = |Y - \hat{\mu}(\X)|$. Let $t_{1-\alpha}^*(\x)$ be $(1-\alpha)$-quantile of the regression residual, as in Definition~\ref{def:oracle_PI}. Then, 
    \begin{enumerate}
        \item If $\x_1, \x_2 \in A$, then $t_{1-\alpha}^*(\x_1) = t_{1-\alpha}^*(\x_2)$ for every $\alpha \in (0,1)$ \label{item_1_eq_relation}
        \item\label{item_2_eq_relation} If $\mathcal{J}$ is another partition of $\sX$ such that, for any $J \in \mathcal{J}$, $\x_1, \x_2 \in J$ implies that $t_{1-\alpha}^*(\x_1) = t_{1-\alpha}^*(\x_2)$ for every $\alpha \in (0,1)$, then 
        \begin{equation*}
          \x_1, \x_2 \in J \implies \x_1, \x_2 \in A.  
        \end{equation*}
    \end{enumerate}
\end{Theorem}
\locart{} attempts to recover the partition described in this theorem.  We explore this approach in the next section.

\subsection{\locart{}: Local Coverage Regression Trees}
\label{sec:locart}



A
regression tree naturally induces a partition of the feature space. Moreover,
 as discussed in Section \ref{sec:regression_trees}, regression trees are consistent estimators of conditional distributions.  
 Thus, a regression tree that predicts the residual $\hat s(\X,Y)$ using $\x$ as an input attempts to recover the partition described by Theorem \ref{thm:optimal_cutoff_eq_relation}, that is, the partition that groups features according to the conditional distribution of the residuals. This is, therefore, how \locart{} builds
 the partition $\sA$. We detail \locart's algorithm and practical considerations in Section \ref{subsec:locart_algorithm} and formalize its theoretical properties in Section \ref{subsec:locart_properties}.



\subsubsection{The algorithm}
\label{subsec:locart_algorithm}

Given a regression estimator $\hat{\mu}(\cdot)$ trained on $\{ (\X_i, Y_i) \}_{i \in I_{\train}} $ and the calibration split with indices $I_{\cal}$, \locart{}  consists of four steps, as outlined in~\ref{fig:locart}:
\begin{enumerate}
    \item Obtain the regression residuals $\hat{s}_i$ of $\hat{\mu}$ on $I_{\cal}$, that is, compute $\hat{s}_i = \hat{s}(\X_i, Y_i) = |\hat{\mu}(\X_i) - Y_i|$, for every $i \in I_{\cal}$.

    \item \label{step_2_locart} Create the dataset $\{(\X_i, \hat{s}_i) \}_{i \in I_{\cal}}$ of pairs of features and regression residuals computed on the calibration dataset, and split it into two disjoint sets $I_{\part}$ and $I_{\cut}$.

    \item Fit a regression tree on $\{(\X_i, \hat{s}_i) \}_{i \in I_{\text{part}}}$ that predicts the residuals $\hat{s}$ based on the features $\X$. This tree induces a partition $\sA$ of the feature space generated by its terminal nodes, as discussed in Section~\ref{sec:preliminaries}. Let $\mT: \sX \to \sA$ be the function mapping an element of $\sX$ to the partition element it belongs.
    
    \label{step_3_locart}

    \item \label{step_4_locart} Estimate $\hat{t}_{1-\alpha}(\x_{n+1})$ for a new instance $\x_{n+1}$ by applying the conformal approach to all instances in $\{(\X_i,s_i) \}_{i \in I_{\text{cut}}}$ that fall into the same element of $\sA$  as $\x_{n+1}$ does. That is, we estimate $\hat{t}_{1-\alpha}(\x_{n + 1})$ as
\begin{align}
     \hat{t}_{1-\alpha}(\x_{n + 1}) = \hat{q}_{1-\alpha}( A(\x_{n+1}))
\label{eq:cutoff-locart}     
\end{align}
where 
\begin{align}
\label{eq:locart_partition_func}
     A(\x_{n+1}) = \left\{ \hat{s}_i : \mT(\x_i) = \mT(\x_{n+1}), (\X_i, \hat{s}_i) \in \{(\X_i,\hat{s}_i) \}_{i \in I_{\text{cut}}}  \right\}.
 \end{align}
 \item \label{step_5_locart} Using  $\hat{t}_{1-\alpha}(\x_{n + 1})$ from Equation~\ref{eq:cutoff-locart}, define the \locart{} prediction interval as $$C_{\locart{}}(\x) = (\hat{\mu}(\x) - \hat{t}_{1 - \alpha}(\x), \hat{\mu}(\x) + \hat{t}_{1 - \alpha}(\x)).$$ 
\end{enumerate}

\definecolor{locartgreen}{RGB}{0, 128, 0}
\definecolor{locartblue}{RGB}{0, 67, 128}

\begin{figure}[!t]
    \centering
    \includegraphics{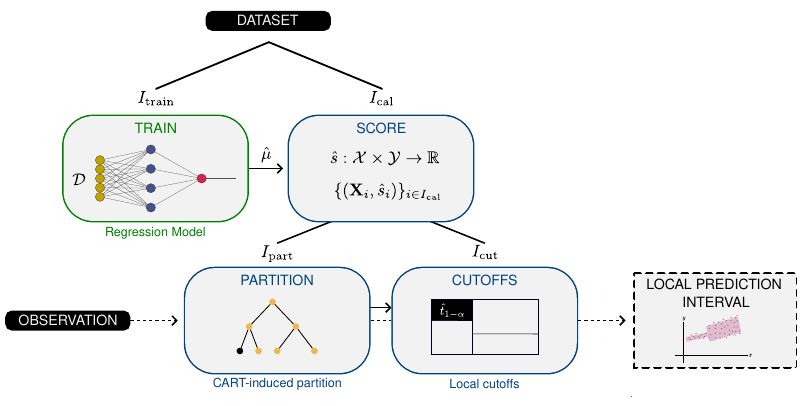}
    \caption{Schematic diagram of \locart{}. \textcolor{locartgreen}{Train}: a regression model (e.g., neural network) is fitted on the $I_{\train}$ split, producing the estimator $\hat{\mu}$. \textcolor{locartblue}{Score}: the regression residuals are calculated on $I_{\cal}$ and a new dataset of features and residuals produced. This dataset is split into $I_{\part}$, used to create a partition of the feature space by fitting a regression tree of residuals, and $I_{\cut}$, used to populate this partition and calculate cutoffs that calibrate prediction intervals locally. In each partition element, the residuals are (approximately) uniformly distributed. As a result, the local thresholds computed in each leaf will converge to the oracle threshold in the asymptotic regime. The path of a new observation (\protect \tikz[baseline = -0.5ex] \protect \draw[line width=1pt,dash pattern=on 2pt off 2pt] (0,0) -- (0.3,0);): the partition element it belongs to is identified; then, the local cutoff is retrieved; next, this cutoff is used to create an adaptive interval centered at the prediction of the regression model for this observation.}
    \label{fig:locart}
\end{figure}

 When using the regression tree algorithm to partition, we may need to control tree growth to avoid empty or redundant partitions. To do that, we perform pre and post-pruning. The pre-pruning is done by fixing the \texttt{min\_samples\_split} hyperparameter in values such as 100 samples, which enables us to obtain well-populated partitions/leaves resulting in accurate cutoff estimations. For the post-pruning step, we remove extra leaves and nodes via cost-complexity pruning (as implemented in the \textit{scikit-learn} library) until we balance partition complexity and predictive performance, reducing the amount of less informative partition elements. 

The remaining regression tree hyperparameters are fixed at \textit{scikit-learn}'s defaults. As a result, we can construct our flexible and adaptive partitions without hyperparameter tuning, which requires multiple refits and could drastically increase the algorithm's running time. Plus, as we make use of the \textit{scikit-learn}'s efficient CART implementation, \locart{} often outperforms benchmarks in the run-time comparison (see Section \ref{sec:experiments}). It is also worth noticing that, to increase \locart{}'s power and circumvent sample size reduction in the calibration step, the splitting of the calibration set presented in Step \ref{step_2_locart} can be skipped, as it is only necessary to derive the theoretical properties explored in Section \ref{subsec:locart_properties}. The empirical results in Section \ref{sec:experiments} reinforce this observation, as \locart{} presents good performances without splitting the calibration set.





\subsubsection{Theoretical properties}
\label{subsec:locart_properties}

By construction of the algorithm, if we fix the partition index set $I_{\part}$, \locart{} produces a fixed partition $\sA$ of the feature space, and thus, satisfies Theorem \ref{theorem:local-validity-splitcp} (i.e., it is locally valid). Consequently, it also satisfies marginal validity. Corollaries \ref{thm:local-Locart} and \ref{thm:marginal-Locart} formalize both statements:

\begin{Corollary}[Local coverage property of \locart{}]
\label{thm:local-Locart} 
Given an i.i.d. sequence $(\X_i, Y_i)_{i=1}^{n+1}$ and a fixed partition $\sA$ of the feature space $\sX$ induced by \locart{} using $I_{\part}$, the predictive intervals of \locart{} satisfy:
\begin{equation*}
    \P[Y_{n+1} \in C_{\locart{}}(\X_{n+1})|\X_{n+1} \in A, I_{\part}, I_{\train}] \ge 1-\alpha,
\end{equation*}
for all $A \in \sA$. 
\end{Corollary}

\begin{Corollary}[Marginal coverage property of \locart{}]
\label{thm:marginal-Locart}
Given an exchangeable sequence $(\X_i, Y_i)_{i=1}^{n+1}$, the predictive intervals of \locart{} satisfy:
\begin{equation*}
    \P[Y_{n+1} \in C_{\locart{}}(\X_{n+1})|I_{\train}] \ge 1-\alpha.
\end{equation*}
\end{Corollary}



Also, notice that the \locart{} cutoff $\hat{t}_{1 - \alpha}(\x_{n + 1})$ can be written as the inverse of the conditional distribution estimator from Equation \ref{eq:qrf} at level $1 - \alpha$, that is:
 \begin{align} \label{eq:locart_cutoff_qrf}
     \hat{t}_{1 - \alpha}(\X_{n + 1}) = \hat{F}^{-1}(1 - \alpha|\X_{n + 1}) \; .
 \end{align} 
 Thus, it is intuitive to think that, by the consistency of regression trees as conditional distribution estimators \citep{meinshausen2006quantile}, the \locart{} cutoff will be close to the oracle cutoff as sample size increases. Indeed, under additional assumptions about the partition's scheme and conditional cumulative distribution, \locart{} converges to the oracle prediction interval on the asymptotic regime. 
 
 To show this, we first assume that asymptotically: (i) the partition elements' are well-populated and (ii) the partition is sufficiently thin:
 \begin{Assumption} \label{assump:CART_functions}
    Let $A_n(\x)$, denote the \locart{} partition element assigned to $\x$, as in Equation \ref{eq:locart_partition_func}, when $I_{cut}$ has size $n$. Then,
    \begin{align*}
        \lim_{n \rightarrow \infty} \inf_{\x \in \sX} n \cdot \P[A_n(\x)] > 0 .
    \end{align*}
\end{Assumption}
Also, we assume that the partition-based $1 - \alpha$ quantile of the conditional distribution of $Y$ is close to the $1 - \alpha$ quantile of the true conditional distribution of $Y$ given $\X$: 
\begin{Assumption}
\label{assump:continuity}
    Let $F_\x(y|\X_i) \coloneqq \P[Y_i \leq y|\X_i \in A_n(\x)]$. For every $\x$, $F_\x$ is continuous and increasing in a neighborhood around $F^{-1}(1 - \alpha|\x)$ and $\sup_{\x \in \sX}|F^{-1}_{\x}(1 - \alpha|\X_i) - F^{-1}(1 - \alpha|\x)| = o_{\P}(1)$.
\end{Assumption}
Under both assumptions,  \locart{} converges to the oracle prediction interval in Theorem \ref{thm:consistency_locart}:

\begin{Theorem}[Consistency of \locart{}]
    \label{thm:consistency_locart}
    Let  $(\X_i, Y_i)_{i=1}^{n + 1}$ be an i.i.d sequence. Consider $\hat{t}_{1-\alpha}(\X_{n + 1})$ and $t_{1-\alpha}^*(\X_{n + 1})$ the \locart{} cutoff estimate and the oracle cutoff for the test point $\X_{n + 1}$. Then, under Assumptions \ref{assump:CART_functions} and \ref{assump:continuity} , $|t^*(\X_{n + 1}) - \hat{t}_{1-\alpha}(\X_{n + 1})| = o_{\P}(1)$.
\end{Theorem}

Finally, by using Theorem \ref{thm:consistency_locart} in the asymptotic regime, we show that \locart{} satisfies asymptotic conditional validity, as formalized in Theorem \ref{thm:asympt-cond-Locart}.

\begin{Theorem}[Asymptotic conditional validity of \locart{}]
\label{thm:asympt-cond-Locart}
Let $(\X_i, Y_i)_{i = 1}^{n + 1}$ be an i.i.d sequence. Under Assumptions \ref{assump:CART_functions} and \ref{assump:continuity}, \locart{} satisfies asymptotic conditional validity.
\end{Theorem}

\subsection{\loforest{}: Local Coverage Regression Forest}
\label{sec:loforest}

\begin{figure}[h]
    \centering
    \includegraphics{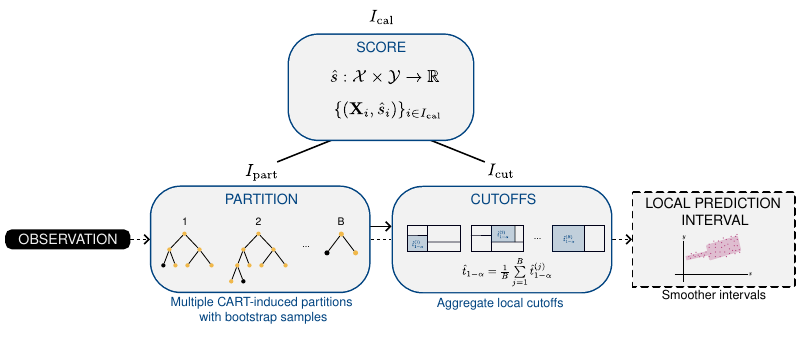}
    \caption{Schematic diagram of \loforest{}, as a modification of the scoring procedure of \locart{}. The regression residuals are calculated on $I_{\cal}$, and a new dataset of features and residuals is produced. This dataset is split into $I_{\part}$, used to create multiple partitions (each one associated with a bootstrap sample of $I_{\part}$) of the feature space by fitting a Random Forest of residuals, and $I_{\cut}$, used to populate these partitions and calculate the cutoffs for each tree. The path of a new observation (\protect\tikz[baseline=-0.5ex] \protect \draw[line width=1pt,dash pattern=on 2pt off 2pt] (0,0) -- (0.3,0);): all the partition elements it belongs to are identified; then, the local cutoffs for each partition are retrieved; next, an average of all the cutoffs is used to create an adaptive interval centered at the prediction of the regression model for this observation. As a result of the averaging step, these intervals are smoother than their \locart{} counterparts.}
    \label{fig:loforest_scheme}
\end{figure}

Regression trees have difficulty in modeling 
 some functional relationships such as additive structures \citep{hastie2009elements}. 
Thus, \locart{} may yield suboptimal results in datasets where regression residuals exhibit such behavior. Moreover, given the decision tree partition structure, \locart{} may also provide poor cutoff estimates for instances in the boundaries of its partition elements. 
 \loforest{} improves \locart{} by 
drawing upon the success of Random Forests. It seeks to increase the expressivity of \locart{} and provides smoother cutoff estimates. Concretely, \loforest{}
builds several partitions $\sA_k$ by fitting many decision trees to bootstrap samples (one for each tree) of the original residuals and features in the calibration set. 
Let $B$ be the number of created decision trees. \loforest{}, as depicted in Figure~\ref{fig:loforest_scheme}, consists of five steps:
\begin{enumerate}
    \item Obtain the regression residuals $\hat{s}_i$ of $\hat{\mu}$ on $I_{\cal}$, that is, compute $\hat{s}_i = \hat{s}(\X_i, Y_i) = |\hat{\mu}(\X_i) - Y_i|$, for every $i \in I_{\cal}$.

    \item \label{step_2_loforest} Create the dataset $\{(\X_i,s_i) \}_{i \in I_{\cal}}$ of pairs of features and regression residuals computed on the calibration dataset, and randomly split it into two disjoint sets $I_{\part}$ and $I_{\cut}$.

    \item Fit a Random Forest on $\{(X_i, s_i)\}_{i \in I_{part}}$, which predicts the residuals $\hat{s}$ based on the features $\X$. This algorithm induces several partitions of the feature space $\sA_{j}$, $j=1, \dots, B$, using the decision tree algorithm on each of the $B$ bootstrap samples. \label{item_3_loforest}

    \item For a new instance $\x_{n + 1}$, estimate the local cutoff $\hat{t}_{1- \alpha}^{(k)}(\x_{n + 1})$ in each decision tree the same way as done in the \locart{} algorithm, using the dataset $\{(\X_i, s_i)\}_{i \in I_{cut}}$. \label{item_4_loforest}

    \item \label{item_5_loforest} Compute the final cutoff $\hat{t}_{1 - \alpha}(\x_{n + 1})$ by averaging all the cutoffs obtained in each decision-tree partition $\sA_k$:
    \begin{align}
    \widehat t_{1 - \alpha}(\x_{n+1}) := \frac{1}{n} \sum_{k = 1}^{B} \widehat{t}_{1 - \alpha}^{(k)}(\x_{n + 1}). \; 
    \label{eq:cutoff-loforest}
    \end{align} 
    
    \item Using $t_{1 - \alpha}(\x_{n+1})$ from Equation~\ref{eq:cutoff-loforest}, define the \loforest{} prediction interval as $$C_{\loforest{}}(\x) = (\hat{\mu} -  \widehat t_{1 - \alpha}(\x), \hat{\mu} +  \widehat t_{1 - \alpha}(\x)).$$
\end{enumerate}

As in \locart, to obtain meaningful local cutoffs in each decision tree, it is essential to control each decision tree growth by either pre or post-pruning. Nevertheless, post-pruning can become computationally expensive if applied to all bootstrap trees. Most importantly, it can diminish the variability between the decision trees and undermine the benefit of ensembling different partitions induced by the Random Forest ensemble. 
Thus, for \loforest{}, we only apply a pre-pruning step in the same way as in Section~\ref{subsec:locart_algorithm}, fixing the \texttt{min\_samples\_split} hyperparameter in values such as 100 samples. For the remaining Random Forest hyperparameters, we set the number of trees as 100. 

Our implementation of  \loforest{} uses the \textit{scikit-learn} Random Forest implementation and, as a consequence, is scalable, as shown in the benchmarking studies in Section~\ref{sec:experiments}. Similarly to \locart{}, empirical results attest that we can skip splitting the calibration set presented in step \ref{step_2_loforest} without harming the method's performance. The results presented in Section \ref{sec:experiments} show that \loforest{}  outperforms several baselines with respect to conditional and marginal coverage of its intervals in benchmark datasets.

\subsection{\alocart{} and \aloforest{}: augmenting the feature space}
\label{sec:a-locart}

In Sections \ref{sec:locart} and \ref{sec:loforest}, we discuss how \locart{} and \loforest{} obtain prediction intervals by partitioning the feature space. Such strategies aim to recover the optimal partition from Theorem \ref{thm:optimal_cutoff_eq_relation}, which groups each sample according to the conditional distribution of residuals. 
In both methods, we only use the original features to create one or several partitions of the feature space. 

The idea of the augmented versions of these methods, \alocart{} and \aloforest{}, is to improve the partitioning by providing additional statistics (i.e., other functions of the features) as inputs of the trees. One possible way of doing this is by giving first-order proxies to the conditional distribution of residuals.
This can be done by using an estimate of $\V[Y|\X]$, for example. Notice that using $\widehat \V[Y|\X]$  as a feature is closely connected to the approach adopted in Mondrian Conformal Regressors \citep{Vovk2005, bostrom2020mondrian}. In fact, by using conditional variance estimates as new covariates, we generalize the Mondrian taxonomy by expanding the binning based on $\V[Y|\X]$ to a general partitioning of both $\V[Y|\X]$ and $\sX$. This avoids the need to tune or fix the number of bins, as we use \locart{} and \loforest{} engines to obtain the partitions. Moreover, it also avoids the adversarial setting for the vanilla Mondrian approach where the variance is constant, but the conditional quantiles of residuals are not (as discussed in Section~\ref{sec:intro} and expanded in~\ref{appendix:baselines}).

In general, we may also add as features of our model other proxies for the distribution of the residuals (e.g., their conditional mean absolute deviation, as used by \citet{Lei2018}) or new feature representations (e.g., random Fourier features \citep{rahimi2007random} and random projections \citep{fradkin2003experiments}). Section \ref{sec:experiments} illustrates how the augmentation procedure can improve \locart{} and \loforest{} partitioning.

\subsection{Extensions to other conformal scores and \wloforest{}}

In the previous sections, we focused on partitioning $\sX$ based on the regression residuals. However, as introduced in Section \ref{subsec:conformal_pred}, this is a particular choice of conformity score. In some cases, it is preferable to build locally valid prediction intervals based on other conformity scores, such as the quantile score and the weighted regression score, mainly when performing inferential tasks other than regression (e.g., quantile regression, conditional distribution estimation). 

The \locart{} and \loforest{} frameworks can be straightforwardly generalized to accommodate any of these situations since these methods are completely agnostic to the choice of conformity score. Even the theoretical guarantees for \locart{} are not directly tied to the regression residual and naturally extend to other conformity scores. In Section \ref{sec:experiments}, we showcase the benefits of using the weighted regression score of \citet{Lei2018} 
within \loforest{}, an approach we call \wloforest{}.

\section{Experiments}
\label{sec:experiments}

In this section, we compare the conditional coverage performance of \locart{}, \alocart{}, \loforest{}, \aloforest{} and \wloforest{} to other state-of-the-art methods for conformal regression and regression intervals on simulated and real-world datasets. 

We use Random Forests as the base model for estimating $\hat{\mu}(\cdot)$ across all methods\footnote{Observe that this choice has no relation whatsoever to how \locart{} or \loforest{} is constructed: any other predictive model could have been chosen (e.g., ridge regression, Lasso, neural network).}. We set \texttt{n\_estimators} to 100, i.e., predictions are averaged over 100 trees. We set a common miscoverage level for predictive intervals at $ \alpha = 0.1$. We compare our methods against the following baselines:

\begin{itemize}
    \item Regression split \citep{Vovk2005, Lei2018}: implemented in the \textit{nonconformist} library \citep{nonconformist}. The only hyperparameter is the miscoverage level.
    \item Weighted regression split \citep{Lei2018}: implemented in our library, \href{https://github.com/Monoxido45/clover/tree/main}{\texttt{clover}}. We use a Random Forest regressor to estimate the mean absolute deviation of the residuals. The only hyperparameter is the miscoverage level.
    \item Mondrian split \citep{bostrom2020mondrian}: implemented in our library, \hyperlink{https://github.com/Monoxido45/clover/tree/main}{\texttt{clover}}. We use the default difficulty estimator (variance of predictions across trees in the Random Forest). Difficulty estimates were binned in $k = 30$ groups according to the quantiles $(1/k, 2/k, \dots, \frac{k-1}{k}, 1)$.
    \item QRF-TC \citep{amoukou2023adaptive, guan2021localized}: implemented in the \textit{acpi} library \citep{acpi}. We set the calibration model as a Random Forest Regressor with 100 trees, a minimal node size of 10, a maximum tree depth of 15, and squared error as the split criterion.
\end{itemize}

To evaluate the conditional validity of intervals output by all methods, we employed the SMIS metric \citep{gneiting2007strictly} for real-world datasets and the conditional coverage absolute deviation for synthetic datasets. We also evaluated marginal validity by computing the average marginal coverage for all datasets. These metrics are briefly described in Section \ref{subsec:metrics}. We replicate the experiments 100 times and compute the average and standard error at the end to estimate the various measures of interest. To give a comprehensive analysis, we divide our evaluation into three parts: conformal methods only, non-conformal methods only, and a combined assessment of all methods. Additionally, we assess the running time of each method to compare their scalability.

\newcommand{\test}{\textrm{test}}
\newcommand{\ccad}{\textrm{CCAD}}
\newcommand{\smis}{\textrm{SMIS}}

\subsection{Metrics for assessing marginal and conditional validity}
\label{subsec:metrics}
In this section, we introduce the metrics used in the experiments. Here, we denote by $C(\cdot)$ any prediction interval and by $\{(\x_i, y_i)\}_{i \in I_{\test}}$ a testing set used to compute coverage diagnostics for all compared methods.

\subsubsection{Conditional coverage absolute deviation (CCAD)}

In the simulation study, we know the ground truth distribution of $Y|\X$. Thus, we can directly measure the conditional coverage. We do this by evaluating the conditional coverage absolute deviation (CCAD): for each $i \in I_{\test}$, we obtain a sample $\mathcal{D} = \{(\x_i, Y_1, \dots, \x_i, Y_{B_y}\}$ of size $B_y$ (fixed as $1,000$) from $Y|\X=\x_i$ and compute
\begin{equation*}
     \delta_{\alpha}(\x_i) = \frac{1}{B_y} \sum_{y \in \mathcal{D}} \mathbbm{1}\{y \in C(\x_i)\}.
 \end{equation*}
 
 $\delta_{\alpha}(\x_i)$ is an estimate of the conditional coverage $\P[Y \in C(\X)|\X = \x_i]$. We derive the conditional coverage absolute deviation (CCAD) by averaging the absolute difference between $\delta_{\alpha}(\x_i)$ and the nominal level $1 - \alpha$ for all $i \in I_{\test}$,
\begin{equation}
     \ccad = \frac{1}{|I_{\test}|} \sum_{i \in I_{\test}} \left| \delta_\alpha(\x_i) - (1 - \alpha) \right|.
\end{equation}
    

\subsubsection{Standard Mean Interval Score (SMIS)}


In real-world datasets, we cannot compute CCAD because the data-generating process is unknown. Instead, we use the standard mean interval score (SMIS) from \citet{gneiting2007strictly}. The interval score associated with $y$ and the miscoverage level $\alpha$ is calculated as
\begin{align}
     \text{IS}_{\alpha}(C(\x), y) =&\, (\max{C(\x)} - \min{C(\x))} \nonumber \\
     &+ \frac{2}{\alpha}(\min{C(\x)} - y) \mathbbm{1}\{ y < \min{C(\x)}\} \nonumber \\
     &+  \frac{2}{\alpha}(y - \max{C(\x)}) \mathbbm{1}\{y > \max{C(\x)}\},
 \end{align}
where $\min{C(\x)}$ and $\max{C(\x)}$ are the minimum and maximum of the (closed) prediction interval $C(\x)$. Intuitively, the interval score penalizes large prediction intervals (first term) that do not contain $y$ (second and third terms) in proportion to how far $y$  is from the limits of the interval and the stricter the miscoverage level. In other words, this score prioritizes the narrowest valid (i.e., containing $y$) prediction interval. The standard mean interval score is then the average over the whole testing set $\{(\X_i, Y_i)\}_{i \in I_{\test}}$:
\begin{equation}
     \smis_{\alpha}(C, \mathcal{D}) = \frac{1}{|I_{\test}|} \sum_{i \in I_{\test}} \text{IS}_{\alpha}(C(\X_i), Y_i) \; ,
 \end{equation}
 which gives us a general balance of interval length and coverage of $y$ for the prediction interval method $C$. Thus, it can be interpreted as an approximate measure of conditional coverage.
\subsubsection{Average marginal coverage}

We also estimate  the average marginal coverage $\P[Y \in C(\X)]$ using
 \begin{equation}
     \text{AMC} = \frac{1}{|I_{\test}|} \sum_{i \in I_{\test}} \mathbbm{1}\{ y_i \in C(\x_i)\}.
 \end{equation}     

\subsection{Simulated data performances}

We adopt the simulated settings employed in \citet{izbicki2022cd}, incorporating some modifications to the number of noisy variables and introducing two additional scenarios. Let $\mathbf{X} = (X_1, \dots, X_d)$, where $X_i \overset{\text{i.i.d}}{\sim} \text{Unif}(-1.5, 1.5)$, and $d$ and $p$ denotes the total number of features and relevant features respectively. The selected simulated settings are as follows:
\begin{itemize}
    \item \textbf{Homoscedastic}: 
    \begin{equation*}
        Y|\x \sim N\left(  \frac{2}{p} \sum_{i = 1}^{p} x_i, 1 \right)
    \end{equation*}
    
    \item \textbf{Heteroscedastic}: 
    \begin{equation*}
        Y|\x \sim N\left( \frac{2}{p} \sum_{i = 1}^{p} x_i, 0.25 + 2 \left| \frac{1}{p} \sum_{i = 1}^{p} x_i \right| \right)
    \end{equation*}
    
    \item \textbf{Asymmetric}: 
    \begin{equation*}
        Y|\x =\frac{2}{p} \sum_{i = 1}^{p} x_i + \varepsilon,
    \end{equation*}
    where $\varepsilon \sim \text{Gamma} \left(1 + 
    \gamma \left|\frac{\sum_{i = 1}^{p} x_i}{p}\right|,
    1 + \gamma \left|\frac{\sum_{i = 1}^{p} x_i}{p}\right| \right)$, with $\gamma = 0.6$ and $1.5$
    
    \item \textbf{$t$-residuals}:
    \begin{equation*}
        Y|\x = \frac{2}{p} \sum_{i = 1}^{p} x_i + \varepsilon,
    \end{equation*} 
    where $\varepsilon \sim \text{t}_{4}$.
    
    \item \textbf{Non-Correlated Heteroscedastic}: 
    \begin{equation*}
        Y|\x \sim N\left(1, 0.25 + \left| \frac{2}{p} \sum_{i = 1}^{p} x_i \right| \right)
    \end{equation*}
\end{itemize}

To enhance the diversity of simulated scenarios, we consider three values for the number of significant features $p$: $1, 3$, and $5$ and fix $d$ as $20$ for all possible scenarios. For each value of $p$, we generate a total of 10,000 training and calibration samples and 5,000 testing samples.

In each run, we evaluate the performance of various methods by estimating the mean conditional coverage absolute deviation (CCAD) and the marginal coverage (AMC) associated with the predictive intervals. To determine these metrics, we employ a holdout testing set and measure the running time required for the calibration step for each method.

\begin{figure}[H]
    \centering
    \includegraphics[width=\textwidth]{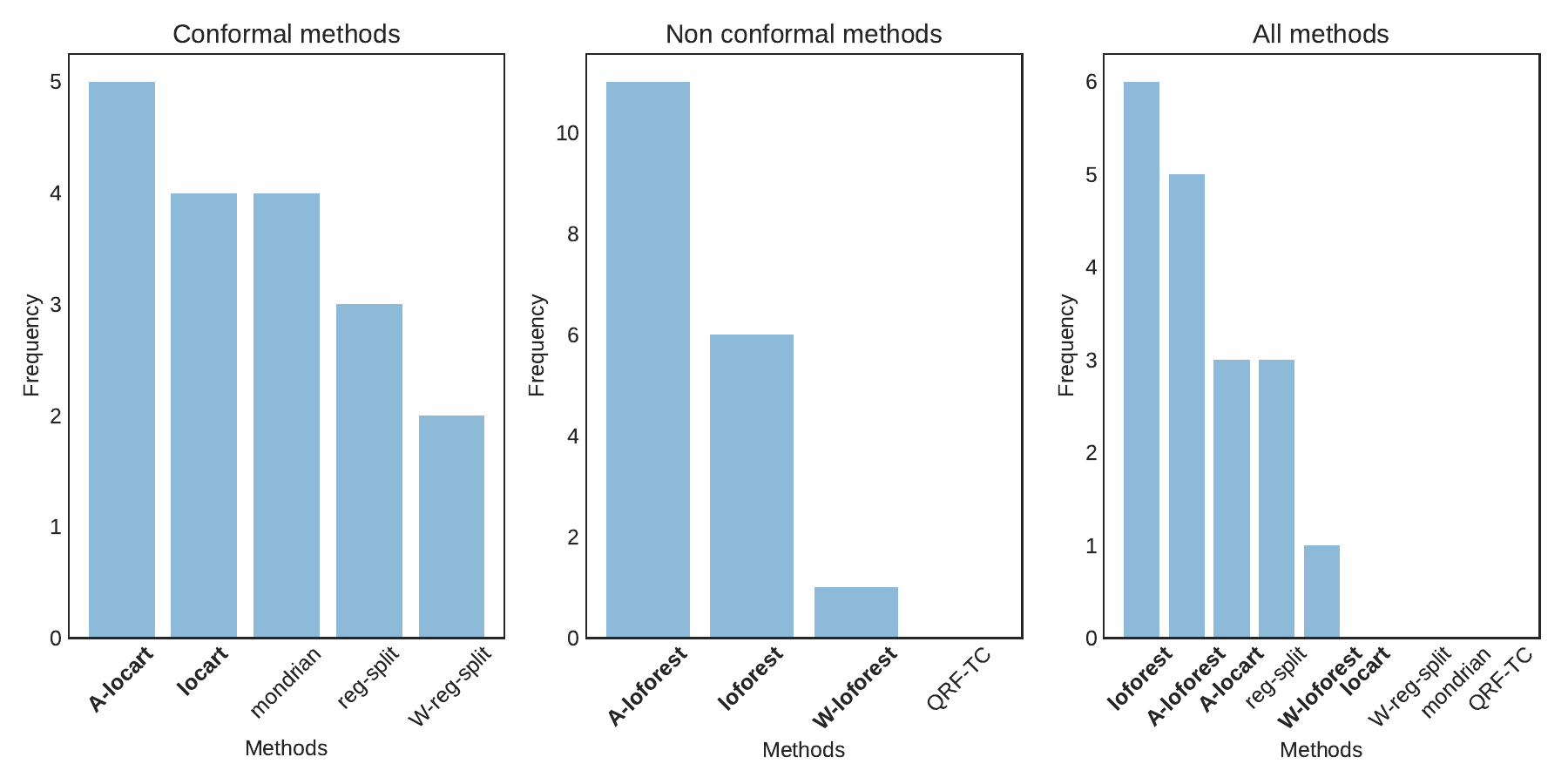}
    \caption{Frequency of times each method performs better in simulated settings according to CCAD in three different comparisons: between only conformal methods (first panel), between only non-conformal methods (second panel), and between all methods altogether (third panel). The plots indicate that  our methods have competitive performance and are flexible to different data scenarios.}
    \label{fig:general_results}
\end{figure}

In general, all methods achieve marginal coverage levels that are very close to the nominal value of 90\%, as shown in Table \ref{tab:marginal_coverage_sim} (see~\ref{sec:appendix_a} for details). However, when it comes to conditional coverage, Table~\ref{tab:conditional_coverage_sim} and Figure~\ref{fig:general_results} indicate variations in performance across different methods:
\begin{itemize}
\item In both homoscedastic settings (homoscedastic and $t$-residuals), we observe in Table \ref{tab:conditional_coverage_sim} that the regression split method demonstrates the best performance, closely followed by \locart{} and \alocart{}. This is because the cutoff $\hat{t}_{1-\alpha}(\x)$ remains constant for $\x$. \locart{} and \alocart{} decision trees capture this behavior and reduce to a trivial tree with only one leaf, effectively replicating the regression split.
\item In all other settings, particularly the asymmetric and heteroscedastic settings, our non-conformal methods (\loforest{}, \aloforest{}) exhibit superior performance. Additionally, among the conformal methods, the first panel in Figure~\ref{fig:general_results} and Table~\ref{tab:conditional_coverage_sim} demonstrate that our two conformal procedures also excel in these settings.
\item QRF-TC does not perform as well as either the other non-conformal or conformal methods, as evident from Table~\ref{tab:conditional_coverage_sim} and the second and third panels in Figure~\ref{fig:general_results}.
\item Considering a comprehensive analysis of all methods and settings, the third panel of Figure~\ref{fig:general_results} indicates that our methods outperform all other regression interval methods.
\end{itemize}

\begin{table}[!http]
\centering
\caption{Mean conditional coverage absolute deviation (CCAD) values for each method and simulation setting. The average across 100 runs is reported with two times the standard error in brackets. Values in bold indicate the methods with better performance in a $95\%$ confidence interval. In general, our methods achieve the best performance in almost all datasets.  
}
\label{tab:conditional_coverage_sim}
\adjustbox{width = \textwidth}{
\begin{tabular}{p{1.9cm}cccccc cccc}
\toprule
\multicolumn{1}{l}{\multirow{3}{*}{Dataset}} & \multicolumn{1}{c}{\multirow{3}{*}{$p$}} & \multicolumn{9}{c}{Type of method} \\ 
\multicolumn{1}{c}{}  & \multicolumn{1}{c}{}  & \multicolumn{5}{c}{Conformal}                                        & \multicolumn{4}{c}{Non-conformal}    \\ \cmidrule(lr){3-7} \cmidrule(lr){8-11}
\multicolumn{1}{c}{}                                    & \multicolumn{1}{c}{}                   & \textbf{\small \locart{}}        & \textbf{\small \alocart{}}      & {\small RegSplit}     & {\small W-RegSplit}  & {\small Mondrian} & \textbf{\small \loforest{}}      & \textbf{\small  \aloforest{}}    & \textbf{\small \wloforest{}}    & {\small QRF-TC}      \\ \midrule
Asym.                     & 1 & \begin{tabular}[c]{@{}c@{}}0.033\\ (0.0004)\end{tabular}          & \begin{tabular}[c]{@{}c@{}}0.036\\ (0.0004)\end{tabular}          & \begin{tabular}[c]{@{}c@{}}0.06\\ (0.0003)\end{tabular}           & \begin{tabular}[c]{@{}c@{}}0.042\\ (0.0002)\end{tabular} & \begin{tabular}[c]{@{}c@{}}0.039\\ (0.0002)\end{tabular} & \begin{tabular}[c]{@{}c@{}}0.028\\ (0.0002)\end{tabular}          & \textbf{\begin{tabular}[c]{@{}c@{}}0.026\\ (0.0002)\end{tabular}} & \begin{tabular}[c]{@{}c@{}}0.039\\ (0.0002)\end{tabular}          & \begin{tabular}[c]{@{}c@{}}0.041\\ (0.0003)\end{tabular} \\
Asym.                     & 3 & \begin{tabular}[c]{@{}c@{}}0.046\\ (0.0007)\end{tabular}          & \begin{tabular}[c]{@{}c@{}}0.04\\ (0.0003)\end{tabular}           & \begin{tabular}[c]{@{}c@{}}0.051\\ (0.0003)\end{tabular}          & \begin{tabular}[c]{@{}c@{}}0.045\\ (0.0003)\end{tabular} & \begin{tabular}[c]{@{}c@{}}0.041\\ (0.0003)\end{tabular} & \begin{tabular}[c]{@{}c@{}}0.037\\ (0.0003)\end{tabular}          & \textbf{\begin{tabular}[c]{@{}c@{}}0.036\\ (0.0002)\end{tabular}} & \begin{tabular}[c]{@{}c@{}}0.041\\ (0.0003)\end{tabular}          & \begin{tabular}[c]{@{}c@{}}0.049\\ (0.0005)\end{tabular} \\
Asym.                     & 5 & \begin{tabular}[c]{@{}c@{}}0.045\\ (0.0003)\end{tabular}          & \begin{tabular}[c]{@{}c@{}}0.04\\ (0.0003)\end{tabular}           & \begin{tabular}[c]{@{}c@{}}0.045\\ (0.0003)\end{tabular}          & \begin{tabular}[c]{@{}c@{}}0.046\\ (0.0003)\end{tabular} & \begin{tabular}[c]{@{}c@{}}0.041\\ (0.0002)\end{tabular} & \begin{tabular}[c]{@{}c@{}}0.039\\ (0.0002)\end{tabular}          & \textbf{\begin{tabular}[c]{@{}c@{}}0.037\\ (0.0002)\end{tabular}} & \begin{tabular}[c]{@{}c@{}}0.043\\ (0.0003)\end{tabular}          & \begin{tabular}[c]{@{}c@{}}0.048\\ (0.0006)\end{tabular} \\ \midrule
Asym. 2                   & 1 & \begin{tabular}[c]{@{}c@{}}0.04\\ (0.0005)\end{tabular}           & \begin{tabular}[c]{@{}c@{}}0.043\\ (0.0005)\end{tabular}          & \begin{tabular}[c]{@{}c@{}}0.086\\ (0.0003)\end{tabular}          & \begin{tabular}[c]{@{}c@{}}0.045\\ (0.0003)\end{tabular} & \begin{tabular}[c]{@{}c@{}}0.041\\ (0.0002)\end{tabular} & \textbf{\begin{tabular}[c]{@{}c@{}}0.027\\ (0.0002)\end{tabular}} & \begin{tabular}[c]{@{}c@{}}0.028\\ (0.0002)\end{tabular}          & \begin{tabular}[c]{@{}c@{}}0.042\\ (0.0002)\end{tabular}          & \begin{tabular}[c]{@{}c@{}}0.049\\ (0.0004)\end{tabular} \\
Asym. 2                   & 3 & \begin{tabular}[c]{@{}c@{}}0.064\\ (0.0007)\end{tabular}          & \begin{tabular}[c]{@{}c@{}}0.055\\ (0.0003)\end{tabular}          & \begin{tabular}[c]{@{}c@{}}0.084\\ (0.0003)\end{tabular}          & \begin{tabular}[c]{@{}c@{}}0.055\\ (0.0003)\end{tabular} & \begin{tabular}[c]{@{}c@{}}0.053\\ (0.0003)\end{tabular} & \textbf{\begin{tabular}[c]{@{}c@{}}0.048\\ (0.0003)\end{tabular}} & \begin{tabular}[c]{@{}c@{}}0.05\\ (0.0002)\end{tabular}           & \begin{tabular}[c]{@{}c@{}}0.051\\ (0.0003)\end{tabular}          & \begin{tabular}[c]{@{}c@{}}0.065\\ (0.0005)\end{tabular} \\
Asym. 2                   & 5 & \begin{tabular}[c]{@{}c@{}}0.072\\ (0.0004)\end{tabular}          & \begin{tabular}[c]{@{}c@{}}0.059\\ (0.0003)\end{tabular}          & \begin{tabular}[c]{@{}c@{}}0.073\\ (0.0003)\end{tabular}          & \begin{tabular}[c]{@{}c@{}}0.063\\ (0.0003)\end{tabular} & \begin{tabular}[c]{@{}c@{}}0.058\\ (0.0002)\end{tabular} & \begin{tabular}[c]{@{}c@{}}0.06\\ (0.0003)\end{tabular}           & \textbf{\begin{tabular}[c]{@{}c@{}}0.054\\ (0.0002)\end{tabular}} & \begin{tabular}[c]{@{}c@{}}0.057\\ (0.0003)\end{tabular}          & \begin{tabular}[c]{@{}c@{}}0.068\\ (0.0005)\end{tabular} \\ \midrule
Heterosc.                & 1 & \begin{tabular}[c]{@{}c@{}}0.029\\ (0.0005)\end{tabular}          & \begin{tabular}[c]{@{}c@{}}0.032\\ (0.0004)\end{tabular}          & \begin{tabular}[c]{@{}c@{}}0.066\\ (0.0002)\end{tabular}          & \begin{tabular}[c]{@{}c@{}}0.041\\ (0.0001)\end{tabular} & \begin{tabular}[c]{@{}c@{}}0.031\\ (0.0002)\end{tabular} & \textbf{\begin{tabular}[c]{@{}c@{}}0.017\\ (0.0002)\end{tabular}} & \begin{tabular}[c]{@{}c@{}}0.019\\ (0.0002)\end{tabular}          & \begin{tabular}[c]{@{}c@{}}0.04\\ (0.0001)\end{tabular}           & \begin{tabular}[c]{@{}c@{}}0.044\\ (0.0004)\end{tabular} \\
Heterosc.                & 3 & \begin{tabular}[c]{@{}c@{}}0.054\\ (0.0006)\end{tabular}          & \begin{tabular}[c]{@{}c@{}}0.046\\ (0.0003)\end{tabular}          & \begin{tabular}[c]{@{}c@{}}0.069\\ (0.0002)\end{tabular}          & \begin{tabular}[c]{@{}c@{}}0.051\\ (0.0002)\end{tabular} & \begin{tabular}[c]{@{}c@{}}0.043\\ (0.0002)\end{tabular} & \textbf{\begin{tabular}[c]{@{}c@{}}0.041\\ (0.0003)\end{tabular}} & \textbf{\begin{tabular}[c]{@{}c@{}}0.041\\ (0.0001)\end{tabular}} & \begin{tabular}[c]{@{}c@{}}0.049\\ (0.0002)\end{tabular}          & \begin{tabular}[c]{@{}c@{}}0.062\\ (0.0004)\end{tabular} \\
Heterosc.                & 5 & \begin{tabular}[c]{@{}c@{}}0.064\\ (0.0004)\end{tabular}          & \begin{tabular}[c]{@{}c@{}}0.052\\ (0.0002)\end{tabular}          & \begin{tabular}[c]{@{}c@{}}0.066\\ (0.0002)\end{tabular}          & \begin{tabular}[c]{@{}c@{}}0.057\\ (0.0002)\end{tabular} & \begin{tabular}[c]{@{}c@{}}0.051\\ (0.0002)\end{tabular} & \begin{tabular}[c]{@{}c@{}}0.056\\ (0.0003)\end{tabular}          & \textbf{\begin{tabular}[c]{@{}c@{}}0.049\\ (0.0002)\end{tabular}} & \begin{tabular}[c]{@{}c@{}}0.054\\ (0.0002)\end{tabular}          & \begin{tabular}[c]{@{}c@{}}0.064\\ (0.0004)\end{tabular} \\ \midrule
Homosc.                 & 1 & \textbf{\begin{tabular}[c]{@{}c@{}}0.01\\ (0.0001)\end{tabular}}  & \textbf{\begin{tabular}[c]{@{}c@{}}0.01\\ (0.0001)\end{tabular}}  & \textbf{\begin{tabular}[c]{@{}c@{}}0.01\\ (0.0001)\end{tabular}}  & \begin{tabular}[c]{@{}c@{}}0.034\\ (0.0002)\end{tabular} & \begin{tabular}[c]{@{}c@{}}0.017\\ (0.0003)\end{tabular} & \begin{tabular}[c]{@{}c@{}}0.012\\ (0.0001)\end{tabular}          & \begin{tabular}[c]{@{}c@{}}0.012\\ (0.0001)\end{tabular}          & \begin{tabular}[c]{@{}c@{}}0.033\\ (0.0001)\end{tabular}          & \begin{tabular}[c]{@{}c@{}}0.028\\ (0.0005)\end{tabular} \\
Homosc.                 & 3 & \textbf{\begin{tabular}[c]{@{}c@{}}0.012\\ (0.0001)\end{tabular}} & \textbf{\begin{tabular}[c]{@{}c@{}}0.012\\ (0.0001)\end{tabular}} & \textbf{\begin{tabular}[c]{@{}c@{}}0.012\\ (0.0001)\end{tabular}} & \begin{tabular}[c]{@{}c@{}}0.035\\ (0.0001)\end{tabular} & \begin{tabular}[c]{@{}c@{}}0.018\\ (0.0003)\end{tabular} & \begin{tabular}[c]{@{}c@{}}0.013\\ (0.0001)\end{tabular}          & \begin{tabular}[c]{@{}c@{}}0.013\\ (0.0001)\end{tabular}          & \begin{tabular}[c]{@{}c@{}}0.033\\ (0.0001)\end{tabular}          & \begin{tabular}[c]{@{}c@{}}0.03\\ (0.0004)\end{tabular}  \\
Homosc.                 & 5 & \textbf{\begin{tabular}[c]{@{}c@{}}0.012\\ (0.0001)\end{tabular}} & \textbf{\begin{tabular}[c]{@{}c@{}}0.012\\ (0.0001)\end{tabular}} & \textbf{\begin{tabular}[c]{@{}c@{}}0.012\\ (0.0001)\end{tabular}} & \begin{tabular}[c]{@{}c@{}}0.035\\ (0.0002)\end{tabular} & \begin{tabular}[c]{@{}c@{}}0.018\\ (0.0002)\end{tabular} & \begin{tabular}[c]{@{}c@{}}0.014\\ (0.0001)\end{tabular}          & \begin{tabular}[c]{@{}c@{}}0.014\\ (0.0001)\end{tabular}          & \begin{tabular}[c]{@{}c@{}}0.033\\ (0.0001)\end{tabular}          & \begin{tabular}[c]{@{}c@{}}0.03\\ (0.0004)\end{tabular}  \\ \midrule
Non-corr. heterosc. & 1 & \begin{tabular}[c]{@{}c@{}}0.029\\ (0.0006)\end{tabular}          & \begin{tabular}[c]{@{}c@{}}0.031\\ (0.0006)\end{tabular}          & \begin{tabular}[c]{@{}c@{}}0.066\\ (0.0002)\end{tabular}          & \begin{tabular}[c]{@{}c@{}}0.041\\ (0.0002)\end{tabular} & \begin{tabular}[c]{@{}c@{}}0.05\\ (0.0003)\end{tabular}  & \textbf{\begin{tabular}[c]{@{}c@{}}0.016\\ (0.0002)\end{tabular}} & \begin{tabular}[c]{@{}c@{}}0.017\\ (0.0002)\end{tabular}          & \begin{tabular}[c]{@{}c@{}}0.04\\ (0.0001)\end{tabular}           & \begin{tabular}[c]{@{}c@{}}0.044\\ (0.0004)\end{tabular} \\
Non-corr. heterosc. & 3 & \begin{tabular}[c]{@{}c@{}}0.054\\ (0.0006)\end{tabular}          & \begin{tabular}[c]{@{}c@{}}0.063\\ (0.0006)\end{tabular}          & \begin{tabular}[c]{@{}c@{}}0.068\\ (0.0002)\end{tabular}          & \begin{tabular}[c]{@{}c@{}}0.05\\ (0.0002)\end{tabular}  & \begin{tabular}[c]{@{}c@{}}0.065\\ (0.0002)\end{tabular} & \textbf{\begin{tabular}[c]{@{}c@{}}0.041\\ (0.0003)\end{tabular}} & \begin{tabular}[c]{@{}c@{}}0.054\\ (0.0004)\end{tabular}          & \begin{tabular}[c]{@{}c@{}}0.048\\ (0.0002)\end{tabular}          & \begin{tabular}[c]{@{}c@{}}0.061\\ (0.0005)\end{tabular} \\
Non-corr. heterosc. & 5 & \begin{tabular}[c]{@{}c@{}}0.064\\ (0.0004)\end{tabular}          & \begin{tabular}[c]{@{}c@{}}0.064\\ (0.0002)\end{tabular}          & \begin{tabular}[c]{@{}c@{}}0.065\\ (0.0002)\end{tabular}          & \begin{tabular}[c]{@{}c@{}}0.057\\ (0.0002)\end{tabular} & \begin{tabular}[c]{@{}c@{}}0.063\\ (0.0002)\end{tabular} & \begin{tabular}[c]{@{}c@{}}0.056\\ (0.0003)\end{tabular}          & \begin{tabular}[c]{@{}c@{}}0.059\\ (0.0002)\end{tabular}          & \textbf{\begin{tabular}[c]{@{}c@{}}0.054\\ (0.0002)\end{tabular}} & \begin{tabular}[c]{@{}c@{}}0.064\\ (0.0004)\end{tabular} \\ \midrule
$t$-residuals                    & 1 & \textbf{\begin{tabular}[c]{@{}c@{}}0.011\\ (0.0001)\end{tabular}} & \textbf{\begin{tabular}[c]{@{}c@{}}0.011\\ (0.0001)\end{tabular}} & \textbf{\begin{tabular}[c]{@{}c@{}}0.011\\ (0.0001)\end{tabular}} & \begin{tabular}[c]{@{}c@{}}0.035\\ (0.0002)\end{tabular} & \begin{tabular}[c]{@{}c@{}}0.017\\ (0.0003)\end{tabular} & \begin{tabular}[c]{@{}c@{}}0.013\\ (0.0001)\end{tabular}          & \begin{tabular}[c]{@{}c@{}}0.013\\ (0.0001)\end{tabular}          & \begin{tabular}[c]{@{}c@{}}0.033\\ (0.0001)\end{tabular}          & \begin{tabular}[c]{@{}c@{}}0.03\\ (0.0006)\end{tabular}  \\
$t$-residuals                    & 3 & \textbf{\begin{tabular}[c]{@{}c@{}}0.012\\ (0.0001)\end{tabular}} & \textbf{\begin{tabular}[c]{@{}c@{}}0.012\\ (0.0001)\end{tabular}} & \textbf{\begin{tabular}[c]{@{}c@{}}0.012\\ (0.0001)\end{tabular}} & \begin{tabular}[c]{@{}c@{}}0.035\\ (0.0002)\end{tabular} & \begin{tabular}[c]{@{}c@{}}0.018\\ (0.0003)\end{tabular} & \begin{tabular}[c]{@{}c@{}}0.014\\ (0.0001)\end{tabular}          & \begin{tabular}[c]{@{}c@{}}0.013\\ (0.0001)\end{tabular}          & \begin{tabular}[c]{@{}c@{}}0.033\\ (0.0001)\end{tabular}          & \begin{tabular}[c]{@{}c@{}}0.03\\ (0.0005)\end{tabular}  \\
$t$-residuals                    & 5 & \textbf{\begin{tabular}[c]{@{}c@{}}0.011\\ (0.0001)\end{tabular}} & \textbf{\begin{tabular}[c]{@{}c@{}}0.011\\ (0.0001)\end{tabular}} & \textbf{\begin{tabular}[c]{@{}c@{}}0.011\\ (0.0001)\end{tabular}} & \begin{tabular}[c]{@{}c@{}}0.035\\ (0.0002)\end{tabular} & \begin{tabular}[c]{@{}c@{}}0.017\\ (0.0003)\end{tabular} & \begin{tabular}[c]{@{}c@{}}0.013\\ (0.0001)\end{tabular}          & \begin{tabular}[c]{@{}c@{}}0.013\\ (0.0001)\end{tabular}          & \begin{tabular}[c]{@{}c@{}}0.032\\ (0.0001)\end{tabular}          & \begin{tabular}[c]{@{}c@{}}0.03\\ (0.0005)\end{tabular}  \\ \bottomrule
\end{tabular}
}
\end{table}

In terms of computational costs, Figure~\ref{fig:running_time_plot} illustrates that \locart{}, \alocart{}, and Mondrian are the most affordable methods, accounting for less than 1\% of the total running time. Except for QRF-TC, the remaining methods exhibit moderate to low computational costs, utilizing between 1\% and 10\% of the total simulation running time. Conversely, QRF-TC stands out as the most resource-intensive method, requiring considerably more than 10\% of the running time to produce predictive intervals. To provide context, when executed on a computer with an Intel Core i7-8700 CPU with 3.20GHz, 6 cores, 12 threads, and 54 GB of RAM, QRF-TC takes an average of approximately 11 minutes to process a dataset comprising 10,000 training and calibration samples with 20 dimensions. In contrast, \wloforest{}, the second most computationally demanding method, halts in approximately 1 minute and 15 seconds.

Therefore, we conclude through these comparisons that, despite some of our methods not being conformal, our framework simultaneously displays excellent conditional validity performance and superior scalability. These results also highlight the flexibility of our methods in adapting to different data-generating processes (e.g., homoscedastic, heteroscedastic, correlated design).


\begin{figure}[H]
    \centering
    \includegraphics[width=\textwidth]{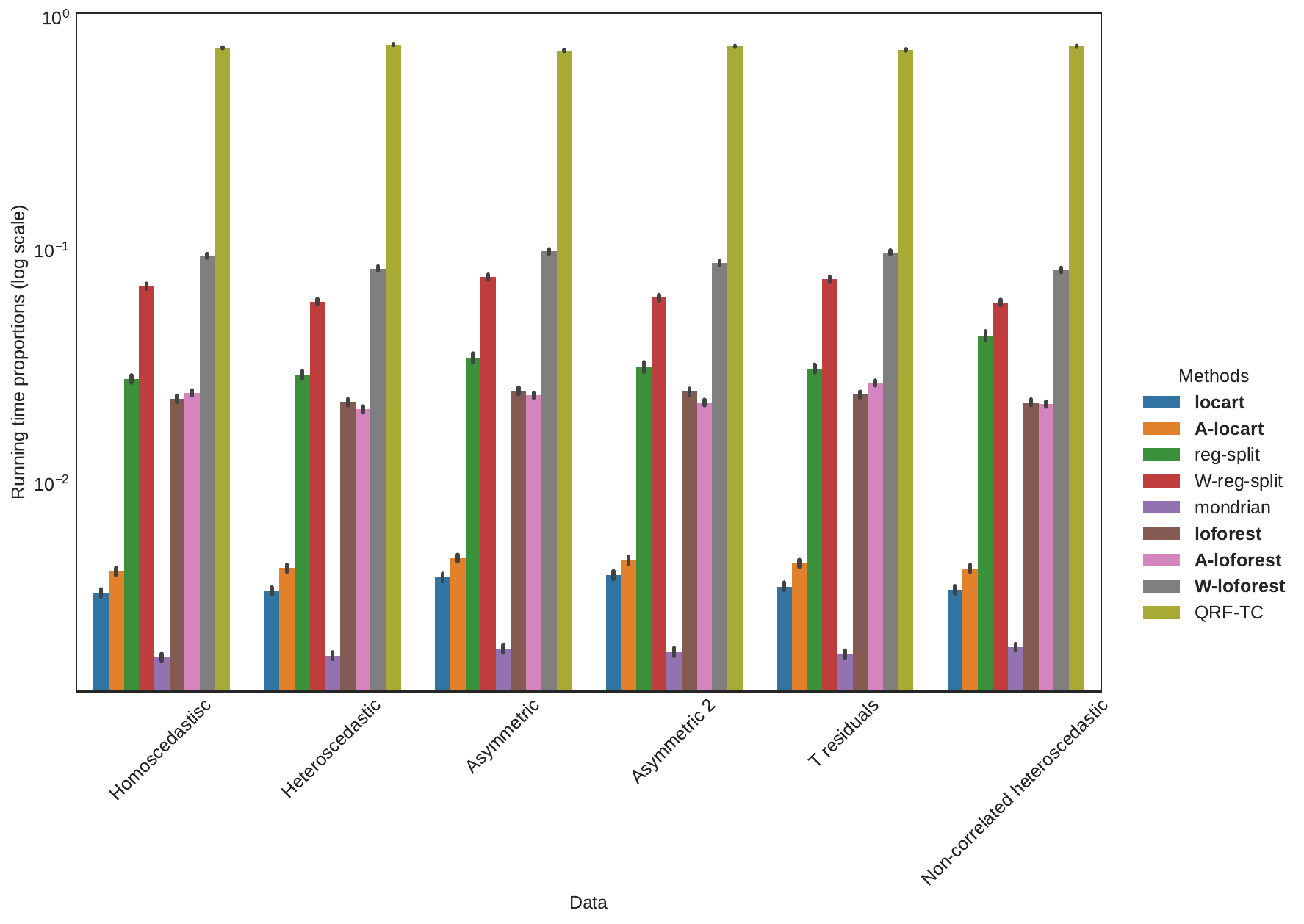}
    \caption{Mean proportion of running time spent in each method for each simulation setting. Our methods are highlighted in bold.  Our methods are scalable compared to competing approaches, and considerably cheaper than LCP-based methods.}
    \label{fig:times}
\end{figure}

\subsection{Real data performances}

Next, we comprehensively compared various regression predictive interval methods using real-world datasets listed in Table \ref{tab:real_data}. Each dataset was split into three distinct sets: training (40\%), calibration (40\%), and testing (20\%). We evaluated the performance of each method by computing the SMIS score, the marginal coverage (AMC) on the testing set, and the running time during the calibration step.

\begin{table}[H]
\centering
\caption{Real-world data information. For each dataset, $n$ is the sample size, and $p$ is the number of features. }
\label{tab:real_data}
\adjustbox{max width=\textwidth }{%
\begin{tabular}{llll @{\hskip 0.5in}  llll}
\toprule 
Dataset & $n$  & $p$ & Source &  Dataset & $n$  & $p$ & Source \\ \midrule                
Concrete   & 1030 & 8 & \href{https://archive.ics.uci.edu/dataset/165/concrete+compressive+strength}{Concrete (UCI)}     &     Electric          & 10000 & 12  & \href{http://archive.ics.uci.edu/ml/datasets/Electrical+Grid+Stability+Simulated+Data+}{Electric (UCI)}              \\
Airfoil    & 1503 & 5   & \href{http://archive.ics.uci.edu/ml/datasets/Airfoil+Self-Noise}{Airfoil (UCI)}                & Bike              & 10886 & 12  & \href{https://www.kaggle.com/code/rajmehra03/bike-sharing-demand-rmsle-0-3194/input?select=train.csv}{Bike (Kaggle)}  \\
Wine white & 1599 & 11  & \href{https://archive.ics.uci.edu/dataset/186/wine+quality}{Wine white (UCI)}                       &     Meps19            & 15781 & 141 & \href{https://github.com/Monoxido45/clover/tree/main/data/raw/meps}{Meps19 (clover repository)}                             \\
Star       & 2161 & 48  & \href{https://github.com/salimamoukou/ACPI/blob/main/acpi/datasets/STAR.csv}{Star (ACPI repository)}  &     Superconductivity & 21263 & 81  & \href{http://archive.ics.uci.edu/ml/datasets/Superconductivty+Data}{Superconductivity (UCI)}                          \\
Wine red   & 4898 & 11  & \href{https://archive.ics.uci.edu/dataset/186/wine+quality}{Wine red (UCI)}                          &     News              & 39644 & 59  & \href{http://archive.ics.uci.edu/dataset/332/online+news+popularity}{News (UCI)}                                      \\
Cycle      & 9568 & 4   & \href{http://archive.ics.uci.edu/dataset/294/combined+cycle+power+plant}{Cycle (UCI)}                &     Protein           & 45730 & 8   & \href{http://archive.ics.uci.edu/dataset/265/physicochemical+properties+of+protein+tertiary+structure}{Protein (UCI)} \\ \bottomrule
\end{tabular}
}
\end{table}

Regarding the marginal coverage of each method, Figure \ref{fig:marginal_coverage_real} in \ref{sec:appendix_a} reveals two main observations:

\begin{itemize}
\item Except for the Mondrian method, all conformal methods demonstrate marginal coverage that is close to the nominal level across all datasets. However, the Mondrian method exhibits over-coverage in small datasets such as winewhite, winered, concrete, and airfoil. This can be attributed to the large number of fixed bins ($k = 30$) relative to the small sample sizes, resulting in bins with only a few instances grouped together. Consequently, larger cutoffs are generated in each partition.
\item Among the non-conformal methods, some show slight over-coverage in specific datasets such as superconductivity, news, meps19, and bike. Notably, QRF-TC demonstrates occasional instances of under-coverage, as observed in the airfoil and star datasets. Thus, even though some of our methods are non-conformal, they generally exhibit marginal coverage at least greater than the nominal level. 
\end{itemize}

\begin{table}[!h]
\caption{SMIS values for each method and real-world dataset. The average across 100 runs is reported with two times the standard deviation in brackets. Values in bold indicate the method with better performance in a 95\% confidence interval. Both \alocart{} and \aloforest{} have good performance.}
\label{tab:smis_performance_real}
\adjustbox{width = \textwidth}{
\begin{tabular}{p{3cm} ccccccccc}
\toprule
& \multicolumn{9}{c}{Type of method}    \\
\multirow{2}{*}{Dataset}  & \multicolumn{5}{c}{Conformal}  & \multicolumn{4}{c}{Non-conformal}    \\ \cmidrule(lr){2-6} \cmidrule(lr){7-10} 
& \textbf{ \locart{}} & \textbf{ \alocart{}} & RegSplit & W-RegSplit   &{\small Mondrian}  & \textbf{ \loforest{}} & \textbf{ \aloforest{}} & \textbf{ \wloforest{}} & QRF-TC                                                             \\ \midrule
\begin{tabular}[l]{@{}l@{}}Concrete\end{tabular}                     & \begin{tabular}[c]{@{}c@{}}27.301 \\ (0.526)\end{tabular}           & \textbf{\begin{tabular}[c]{@{}c@{}}26.196 \\ (0.497)\end{tabular}}  & \begin{tabular}[c]{@{}c@{}}27.729 \\ (0.542)\end{tabular}           & \begin{tabular}[c]{@{}c@{}}28.290 \\ (0.442)\end{tabular}          & \multicolumn{1}{c}{\begin{tabular}[c]{@{}c@{}}28.490 \\ (0.399)\end{tabular}}          & \begin{tabular}[c]{@{}c@{}}27.729 \\ (0.542)\end{tabular}           & \begin{tabular}[c]{@{}c@{}}27.729 \\ (0.542)\end{tabular}           & \begin{tabular}[c]{@{}c@{}}28.290 \\ (0.442)\end{tabular}          & \textbf{\begin{tabular}[c]{@{}c@{}}25.497 \\ (0.487)\end{tabular}} \\[12pt]
\begin{tabular}[c]{@{}l@{}}Airfoil\end{tabular}                      & \begin{tabular}[c]{@{}c@{}}10.395 \\ (0.170)\end{tabular}           & \begin{tabular}[c]{@{}c@{}}10.232 \\ (0.168)\end{tabular}           & \begin{tabular}[c]{@{}c@{}}10.915 \\ (0.194)\end{tabular}           & \begin{tabular}[c]{@{}c@{}}10.140 \\ (0.176)\end{tabular}          & \multicolumn{1}{c}{\begin{tabular}[c]{@{}c@{}}10.587 \\ (0.157)\end{tabular}}          & \begin{tabular}[c]{@{}c@{}}10.341 \\ (0.173)\end{tabular}           & \begin{tabular}[c]{@{}c@{}}10.201 \\ (0.167)\end{tabular}           & \begin{tabular}[c]{@{}c@{}}10.085 \\ (0.176)\end{tabular}          & \textbf{\begin{tabular}[c]{@{}c@{}}9.533 \\ (0.153)\end{tabular}}  \\[12pt]
\begin{tabular}[c]{@{}l@{}}Wine white\end{tabular}                   & \begin{tabular}[c]{@{}c@{}}2.950 \\ (0.0217)\end{tabular}           & \textbf{\begin{tabular}[c]{@{}c@{}}2.884 \\ (0.023)\end{tabular}}   & \begin{tabular}[c]{@{}l@{}}2.977 \\ (0.022)\end{tabular}            & \begin{tabular}[c]{@{}c@{}}3.012 \\ (0.021)\end{tabular}           & \multicolumn{1}{c}{\textbf{\begin{tabular}[c]{@{}c@{}}2.864 \\ (0.021)\end{tabular}}}  & \begin{tabular}[c]{@{}c@{}}2.905 \\ (0.022)\end{tabular}            & \textbf{\begin{tabular}[c]{@{}c@{}}2.853 \\ (0.021)\end{tabular}}   & \begin{tabular}[c]{@{}c@{}}2.996 \\ (0.021)\end{tabular}           & \textbf{\begin{tabular}[c]{@{}c@{}}2.843 \\ (0.021)\end{tabular}} \\[12pt]
\begin{tabular}[c]{@{}l@{}}Star\end{tabular}                         & \textbf{\begin{tabular}[c]{@{}c@{}}973.786 \\ (6.959)\end{tabular}} & \textbf{\begin{tabular}[c]{@{}c@{}}973.614 \\ (7.094)\end{tabular}} & \textbf{\begin{tabular}[c]{@{}c@{}}972.470 \\ (7.053)\end{tabular}} & \begin{tabular}[c]{@{}c@{}}1042.652 \\ (7.450)\end{tabular}        & \multicolumn{1}{c}{\begin{tabular}[c]{@{}c@{}}1007.822 \\ (7.278)\end{tabular}}        & \textbf{\begin{tabular}[c]{@{}c@{}}970.550 \\ (6.959)\end{tabular}} & \textbf{\begin{tabular}[c]{@{}c@{}}970.513 \\ (6.981)\end{tabular}} & \begin{tabular}[c]{@{}c@{}}1039.369 \\ (7.387)\end{tabular}        & \begin{tabular}[c]{@{}c@{}}997.357 \\ (8.134)\end{tabular}        \\[12pt]
\begin{tabular}[c]{@{}l@{}}Wine red\end{tabular}                     & \begin{tabular}[c]{@{}c@{}}2.780 \\ (0.033)\end{tabular}            & \textbf{\begin{tabular}[c]{@{}c@{}}2.750 \\ (0.0307)\end{tabular}}  & \begin{tabular}[c]{@{}c@{}}2.798 \\ (0.033)\end{tabular}            & \begin{tabular}[c]{@{}c@{}}2.874 \\ (0.031)\end{tabular}           & \multicolumn{1}{c}{\begin{tabular}[c]{@{}c@{}}2.843 \\ (0.031)\end{tabular}}           & \textbf{\begin{tabular}[c]{@{}c@{}}2.758 \\ (0.032)\end{tabular}}   & \textbf{\begin{tabular}[c]{@{}c@{}}2.725 \\ (0.031)\end{tabular}}   & \begin{tabular}[c]{@{}c@{}}2.869 \\ (0.030)\end{tabular}           & \textbf{\begin{tabular}[c]{@{}c@{}}2.725\\ (0.034)\end{tabular}}   \\[12pt]
\begin{tabular}[c]{@{}l@{}}Cycle\end{tabular}                        & \begin{tabular}[c]{@{}c@{}}15.850 \\ (0.118)\end{tabular}           & \textbf{\begin{tabular}[c]{@{}c@{}}15.211 \\ (0.120)\end{tabular}}  & \begin{tabular}[c]{@{}c@{}}15.930 \\ (0.121)\end{tabular}           & \begin{tabular}[c]{@{}c@{}}15.480 \\ (0.110)\end{tabular}          & \multicolumn{1}{c}{\textbf{\begin{tabular}[c]{@{}c@{}}15.191 \\ (0.115)\end{tabular}}} & \begin{tabular}[c]{@{}c@{}}15.623 \\ (0.116)\end{tabular}           & \textbf{\begin{tabular}[c]{@{}c@{}}15.087 \\ (0.118)\end{tabular}}  & \begin{tabular}[c]{@{}c@{}}15.386 \\ (0.109)\end{tabular}          & \begin{tabular}[c]{@{}c@{}}15.420 \\ (0.109)\end{tabular}         \\[12pt]
\begin{tabular}[c]{@{}l@{}}Electric $\times \,\,(10^{-2})$\end{tabular} & \begin{tabular}[c]{@{}c@{}}5.360 \\ (0.03)\end{tabular}             & \begin{tabular}[c]{@{}c@{}}5.320 \\ (0.03)\end{tabular}             & \begin{tabular}[c]{@{}c@{}}5.580 \\ (0.03)\end{tabular}             & \begin{tabular}[c]{@{}c@{}}5.050 \\ (0.02)\end{tabular}            & \multicolumn{1}{c}{\begin{tabular}[c]{@{}c@{}}5.340 \\ (0.03)\end{tabular}}            & \begin{tabular}[c]{@{}c@{}}5.220 \\ (0.02)\end{tabular}             & \begin{tabular}[c]{@{}c@{}}5.190 \\ (0.02)\end{tabular}             & \textbf{\begin{tabular}[c]{@{}c@{}}4.990 \\ (0.02)\end{tabular}}   & \begin{tabular}[c]{@{}c@{}}5.170 \\ (0.02)\end{tabular}           \\[12pt]
\begin{tabular}[c]{@{}l@{}}Bike\end{tabular}                        & \begin{tabular}[c]{@{}c@{}}179.017 \\ (1.549)\end{tabular}          & \begin{tabular}[c]{@{}c@{}}162.536 \\ (1.112)\end{tabular}          & \begin{tabular}[c]{@{}c@{}}221.847 \\ (1.832)\end{tabular}          & \begin{tabular}[c]{@{}c@{}}165.679 \\ (1.098)\end{tabular}         & \multicolumn{1}{c}{\begin{tabular}[c]{@{}c@{}}163.245 \\ (1.126)\end{tabular}}         & \begin{tabular}[c]{@{}c@{}}178.467 \\ (1.397)\end{tabular}          & \textbf{\begin{tabular}[c]{@{}c@{}}160.772 \\ (1.105)\end{tabular}} & \begin{tabular}[c]{@{}c@{}}164.572 \\ (1.074)\end{tabular}         & \begin{tabular}[c]{@{}c@{}}171.750 \\ (1.242)\end{tabular}          \\[12pt]
\begin{tabular}[c]{@{}l@{}}Meps19\end{tabular}                      & \begin{tabular}[c]{@{}c@{}}73.316 \\ (0.977)\end{tabular}           & \textbf{\begin{tabular}[c]{@{}c@{}}67.089 \\ (1.068)\end{tabular}}  & \begin{tabular}[c]{@{}c@{}}105.539 \\ (1.191)\end{tabular}          & \textbf{\begin{tabular}[c]{@{}c@{}}66.575 \\ (1.033)\end{tabular}} & \multicolumn{1}{c}{\textbf{\begin{tabular}[c]{@{}c@{}}66.697 \\ (1.105)\end{tabular}}} & \begin{tabular}[c]{@{}c@{}}71.681 \\ (0.957)\end{tabular}           & \textbf{\begin{tabular}[c]{@{}c@{}}66.585 \\ (1.068)\end{tabular}}  & \textbf{\begin{tabular}[c]{@{}c@{}}66.574 \\ (1.005)\end{tabular}} & \begin{tabular}[c]{@{}c@{}}69.372 \\ (0.966)\end{tabular}         \\[12pt]
\begin{tabular}[c]{@{}l@{}}Superconductivity\end{tabular}           & \begin{tabular}[c]{@{}c@{}}41.248 \\ (0.280)\end{tabular}           & \begin{tabular}[c]{@{}c@{}}37.590 \\ (0.223)\end{tabular}           & \begin{tabular}[c]{@{}c@{}}53.095 \\ (0.283)\end{tabular}           & \begin{tabular}[c]{@{}c@{}}41.005 \\ (0.230)\end{tabular}          & \multicolumn{1}{c}{\begin{tabular}[c]{@{}c@{}}39.165 \\ (0.223)\end{tabular}}          & \begin{tabular}[c]{@{}c@{}}39.953 \\ (0.237)\end{tabular}           & \textbf{\begin{tabular}[c]{@{}c@{}}36.571 \\ (0.212)\end{tabular}}  & \begin{tabular}[c]{@{}c@{}}40.147 \\ (0.225)\end{tabular}          & \begin{tabular}[c]{@{}c@{}}37.352 \\ (0.214)\end{tabular}          \\[12pt]
\begin{tabular}[c]{@{}l@{}}News $\times \, (10^{4})$\end{tabular}      & \begin{tabular}[c]{@{}c@{}}3.304 \\ (0.043)\end{tabular}            & \textbf{\begin{tabular}[c]{@{}c@{}}2.988\\ (0.040)\end{tabular}}    & \begin{tabular}[c]{@{}c@{}}3.571\\ (0.043)\end{tabular}             & \textbf{\begin{tabular}[c]{@{}c@{}}2.953\\ (0.040)\end{tabular}}   & \multicolumn{1}{c}{\textbf{\begin{tabular}[c]{@{}c@{}}2.961 \\ (0.042)\end{tabular}}}  & \begin{tabular}[c]{@{}c@{}}3.163\\ (0.041)\end{tabular}             & \textbf{\begin{tabular}[c]{@{}c@{}}2.911\\ (0.040)\end{tabular}}    & \textbf{\begin{tabular}[c]{@{}c@{}}2.966\\ (0.040)\end{tabular}}   & \begin{tabular}[c]{@{}c@{}}3.124 \\ (0.040)\end{tabular}          \\[12pt]
\begin{tabular}[c]{@{}l@{}}Protein\end{tabular}                     & \begin{tabular}[c]{@{}c@{}}16.607 \\ (0.047)\end{tabular}           & \begin{tabular}[c]{@{}c@{}}14.039 \\ (0.048)\end{tabular}           & \begin{tabular}[c]{@{}c@{}}17.583 \\ (0.036)\end{tabular}           & \begin{tabular}[c]{@{}c@{}}14.693 \\ (0.044)\end{tabular}          & \multicolumn{1}{c}{\begin{tabular}[c]{@{}c@{}}14.045 \\ (0.043)\end{tabular}}          & \begin{tabular}[c]{@{}c@{}}16.303 \\ (0.037)\end{tabular}           & \textbf{\begin{tabular}[c]{@{}c@{}}13.812 \\ (0.046)\end{tabular}}  & \begin{tabular}[c]{@{}c@{}}14.612 \\ (0.043)\end{tabular}          & \begin{tabular}[c]{@{}c@{}}15.388 \\ (0.038)\end{tabular}       \\  \bottomrule
\end{tabular}
}
\end{table}

\begin{figure}[!h]
    \centering
    \includegraphics[width=\textwidth]{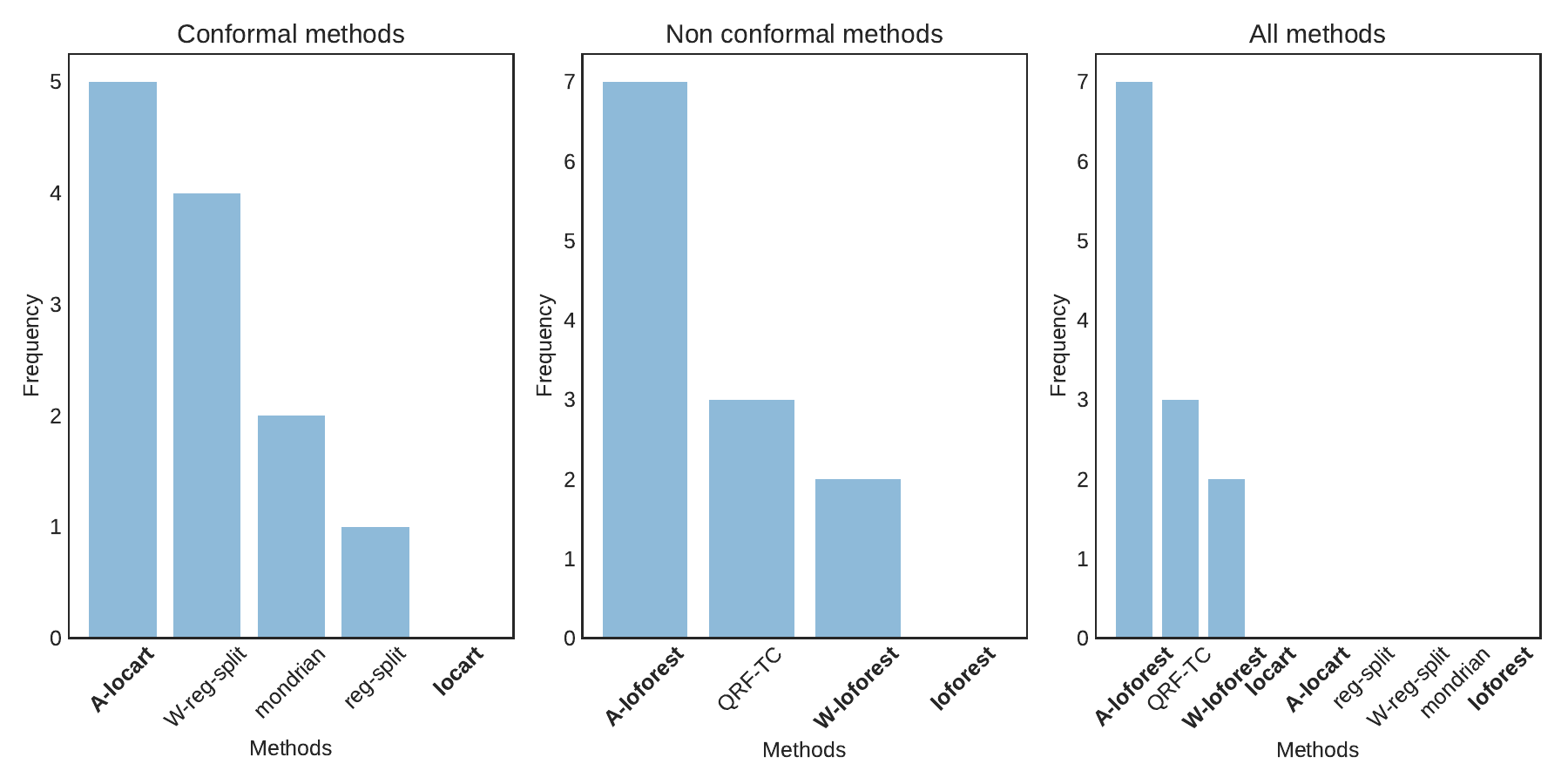}
    \caption{Frequency of times each method performs better in real datasets according to SMIS in three different comparisons: between only conformal methods (first panel), between only non-conformal methods (second panel), and between all methods altogether (third panel). All panels suggest that both \alocart{} and \aloforest{} stand out as the most frequent methods. The third panel suggests also that, among all methods, \aloforest{} has a superior performance across several real datasets.}
    \label{fig:smis_barplot}
\end{figure}

Based on Table~\ref{tab:smis_performance_real} and Figure 
\ref{fig:smis_barplot}, we also conclude the following:
\begin{itemize}
    \item  Table \ref{tab:smis_performance_real} demonstrates that our conformal and non-conformal methods, particularly \aloforest{} and \alocart{}, exhibit strong performance across most datasets, except for the airfoil dataset. Notably, \aloforest{} shows remarkable performance in larger datasets such as protein, superconductivity, and bike-sharing data. This indicates that our adaptive partitioning of the feature space effectively functions in larger data and achieves favorable conditional coverage in such settings.
    \item   Analyzing the first panel of Figure~\ref{fig:smis_barplot}, we observe that our conformal proposal outperforms competing conformal methods in five datasets. Moreover, by referring to Table~\ref{tab:smis_performance_real}, we find that \alocart{} generally performs equally or better than Mondrian, suggesting that our approach enhances the conditional variance binning of Mondrian by incorporating other features and increasing their flexibility. 
    \item Further examination of Table~\ref{tab:smis_performance_real} reveals that QRF-TC achieves superior performance in small datasets, notably the airfoil dataset, but lags behind other non-conformal and conformal methods in all other scenarios.
    \item Notably, for the star dataset, as indicated in Table~\ref{tab:smis_performance_real}, the regression split combined with our set of methods (\locart{}, \alocart{}, \loforest{}, \aloforest{}) outperforms other approaches.
    \item Overall, Figure~\ref{fig:smis_barplot} demonstrates that our class of methods achieves strong conditional coverage performance across real-world datasets.
\end{itemize}

\begin{figure}[http]
    \centering
    \includegraphics[width=\textwidth]{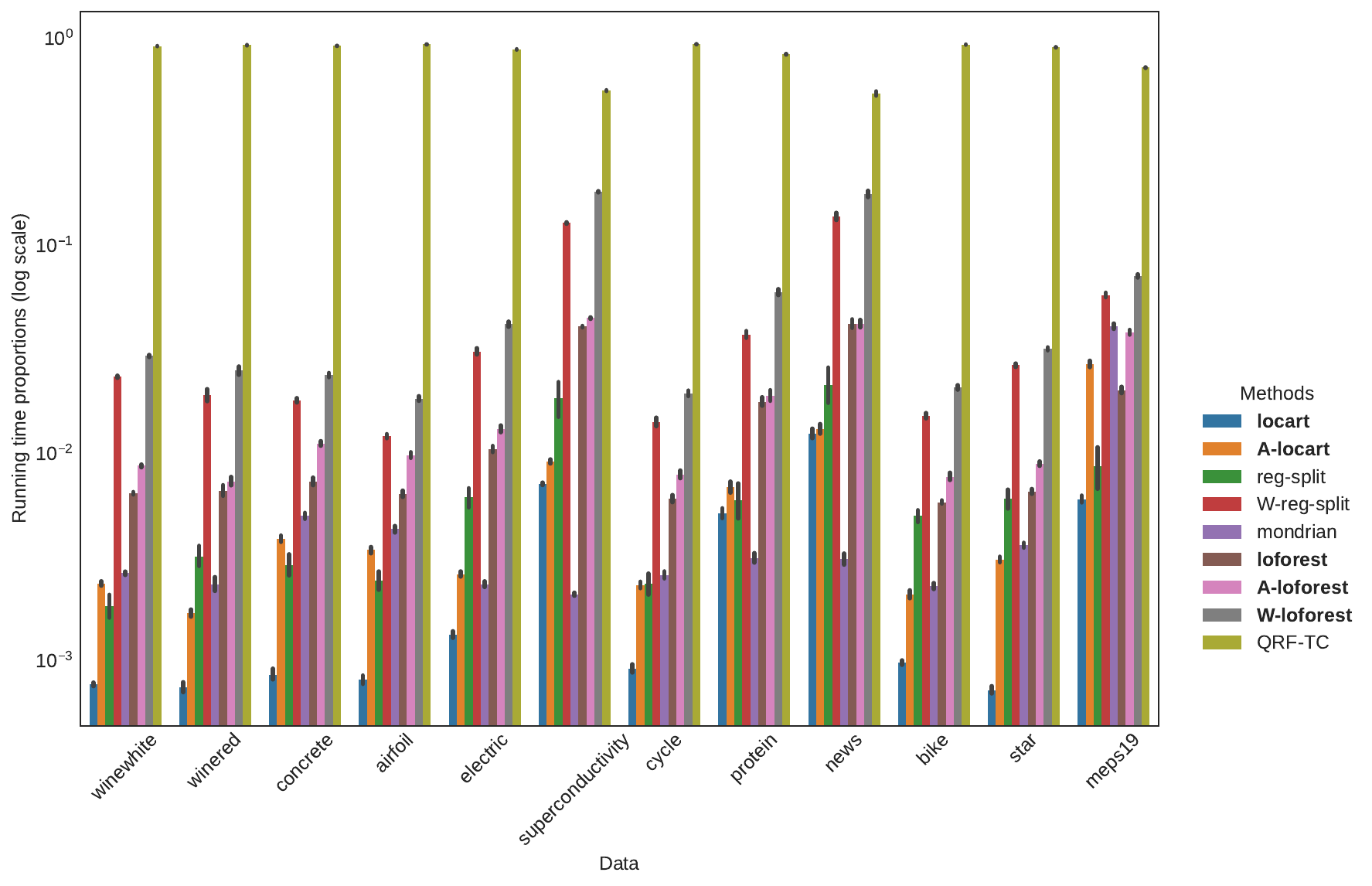}
    \caption{Mean proportion of running time spent in each method for each real dataset. Our methods are highlighted in bold. Even for medium to big data,  our methods are still scalable compared to the remaining approaches.}
    \label{fig:running_time_plot}
\end{figure}

Regarding the computational cost analysis of each method, Figure~\ref{fig:running_time_plot} exhibits a behavior similar to what was observed in the simulated experiments, but with some variations. In addition to \locart{}, \alocart{}, and Mondrian, regression split is also identified as one of the most computationally efficient methods, with a running time spanning from 0.1\% to close to 1\% of the total analysis running time. Although the remaining methods, except for QRF-TC, demonstrate moderate to low computational costs (time consumption between 1\% and 10\%) for the majority of datasets, W-reg-split and \wloforest{} exceed 10\% consumption in the superconductivity and news datasets. This may be attributed to the high computational cost of training the MAD model in these contexts, which is avoided by both \loforest{} and \aloforest{}.

Finally, we reinforce that QRF-TC is considerably more computationally costly than all the other methods in real data settings. For comparison, QRF-TC takes on average 14.4 seconds and 5 minutes to run concrete and protein datasets, respectively, while W-Loforest (the second most computationally heavy algorithm) takes 0.4 and 23 seconds to run each.

\section{Final Remarks}
\label{sec:final}

We propose two methods to calibrate prediction intervals: \locart{} and \loforest{}. These methods produce adaptive prediction intervals based on a sensible partition of the feature space created by training regression trees on conformity scores. A central aspect of \locart{} and \loforest{} is their applicability to high-dimensional settings, as our partition is not based on distance metrics on the feature space. Instead, this partition is obtained by grouping instances according to the conditional distribution of the conformal scores. Furthermore, these methods can be extended to calibrate prediction intervals based on any conformal score.


From a theoretical perspective, we show that \locart{} produces prediction intervals with marginal and local coverage guarantees. In addition, we prove that, under additional assumptions, \locart{} has asymptotic conditional coverage.
Since \loforest{} is essentially an ensemble of \locart{} instances, we expect that \loforest{} inherits the properties of \locart{} while stabilizing the cutoff estimates that define the prediction intervals.

We validated our methods in simulated and real datasets, where they exhibited superior performance compared to baselines. In the simulated settings, our methods and their extensions, such as \alocart{} and \aloforest{}, present good conditional coverage and control the marginal coverage at the nominal level. The same holds when applying these methods to real data for both small and large samples. In all scenarios, our methods outperform established state-of-art algorithms, such as LCP and QRF-TC, including in running time comparisons. 


We provide a Python package, named \href{https://github.com/Monoxido45/clover}{\texttt{clover}}, that implements our methods using the \textit{scikit-learn} interface and replicates all experiments done in this work. For future work, we intend to apply our class of methods to other conformal scores, thus improving their local coverage properties.







\section{Acknowledgments}
L. M. C. C. is grateful for the fellowship provided by Funda\c{c}\~{a}o de Amparo \`{a} Pesquisa do Estado de S\~{a}o Paulo (FAPESP), grant 2022/08579-7.
R. I. is grateful for 
the financial support of FAPESP (grants 2019/11321-9 and 2023/07068-1) and 
CNPq (grant 309607/2020-5). 
M. P. O. was supported through grant 2021/02178-8, São Paulo Research Foundation (FAPESP).
R. B. S. produced this work as
part of the activities of FAPESP Research, Innovation 
and Dissemination Center for Neuromathematics
(grant 2013/07699-0). The authors are also grateful for the suggestions given by Rodrigo F. L. Lassance.

\appendix

\section{Failures of baselines}
\label{appendix:baselines}

We recall, from the discussion of Section~\ref{sec:motivation}, that the oracle prediction interval depends on the conditional quantiles of the regression residual (calculated with the true regression function). Unless the data is homoscedastic, the residual is non-constant across the feature space, and so is its quantile when seen as a function of $\x \in \sX$. Therefore, any method that seeks to approximate the oracle prediction interval (and inherit its optimal properties) should be based on conformity scores that adapt to the data's local features. In the following, we present some data-generating processes where purportedly adaptive methodologies fail. 

\subsection{Locally weighted regression split } \label{subsec:weighted}

Let $\hat{\mu}$ be an estimator of the regression function. \citeauthor{Lei2018} uses the conditional mean absolute deviation (MAD) of $[Y-\hat{\mu}(X)] | X=x $ to normalize prediction intervals. To see why this might fail, consider
\begin{align*}
Y|X = x \sim 
\begin{cases}
\textrm{Laplace}(0, M), &\quad x \in [0, 1), \\
\textrm{Beta}(\alpha, \beta),  &\quad x \in [1, 2],
\end{cases}    
\end{align*}
where $\alpha = 2, \beta = 2$, and $  M = \frac{2 \alpha^\alpha \beta^\beta}{B(\alpha, \beta)(\alpha+\beta)^{\alpha+\beta+1}}$, with $B(\alpha, \beta)$ the Beta function. The conditional mean absolute deviation of $Y| X=x$ is constant and equal to $M$. The conditional quantiles of $|Y-\mu(X)|$, however, depend on whether $x \in [0, 1)$ or $x \in [1, 2]$. Therefore, normalizing with the MAD does not render the locally weighted regression residual adaptive in this situation.

\paragraph{Difficulty as conditional variance of $Y$}\label{par:conditional-var}

\citeauthor{bostrom2017accelerating} uses the empirical variance of predictions made by the trees of a Random Forest to normalize prediction regions. Even in the hypothetical scenario where we replace the empirical variance with the true conditional variance of the labels, $\V(Y|X=x)$, the variance-normalized prediction intervals cannot yield asymptotically valid prediction intervals. To see this, consider the mixture model
\begin{equation*}
    Y|X = x  \sim  0.5\, N(-x, s^2-x^2) + 0.5 \, N(x, s^2-x^2),
\end{equation*}
where we take $x \in [0,1]$ for simplicity and $s$ sufficiently large so as to $Y|X$ be unimodal. It is easy to show that $\V(Y|X=x) = s^2$, thus independent of $x$. On the other hand, the conditional quantiles of $|Y-\mu(X)|$ or of $|Y||X=x$ (since $\mu(x)=  0$) depends on $x$: as $x$ increases, the density is concentrated at larger values of $x$, forcing the quantiles to drift away from zero. It follows that even if the samples have the same conditional variance -- and thus, would be binned together in the approach of \citeauthor{bostrom2017accelerating} -- they do not have the same distribution of regression residuals. Therefore, normalization with conditional variance cannot be used to construct prediction intervals that adapt to the data heteroscedasticity.

\section{Proofs}
\label{appendix:proofs}

The proofs are organized into four subsections: \ref{subsec:proof-local-validity-splicp} proves that \locart{} has local and marginal validity (Theorem \ref{theorem:local-validity-splitcp} and Corollaries \ref{thm:local-Locart} and \ref{thm:marginal-Locart}), \ref{subsec:proof_optimal_partition} proves Theorem \ref{thm:optimal_cutoff_eq_relation}, \ref{subsec:proof_consistency_locart} proves the consistency of \locart{} (Theorem \ref{thm:consistency_locart}), and \ref{subsec:proof_asympt_valid} proves \locart{} asymptotic conditional validity (Theorem \ref{thm:asympt-cond-Locart}). Additionally, to conduct the proof on the last subsection (\ref{subsec:proof_asympt_valid}) we make use of the following definitions:
\begin{Definition}[Convergence to the oracle prediction interval]
\label{def:convergence_to_oracle}
   A prediction interval method $C(\cdot)$ converges to the oracle prediction interval $C^*(\cdot)$ if:
    \begin{align*}
        \P[Y_{n + 1} \in C^*(\X_{n + 1}) \Delta C(\X_{n + 1})] = o(1), \; \text{where $A \Delta B := (A \cap B^c) \cup (B \cap A^c)$} \; .
    \end{align*}
\end{Definition}
\begin{Definition}[Asymptotic conditional validity \citep{Lei2018}]
\label{def:asymp_cond_valid}
    A prediction interval method $C(\cdot)$ satisfies asymptotic conditional validity if there exist random sets $\Lambda_n$, such that $\P[X_{n + 1} \in \Lambda_n| \Lambda_n] = 1 - o_{\P}(1)$ and:
    \begin{align*}
        \sup_{\x_{n + 1} \in \Lambda_n} \left| \P[Y_{n + 1} \in C(\X_{n + 1})| \X_{n + 1} = \x_{n + 1}] - (1 - \alpha) \right| = o(1) \; .
    \end{align*}
\end{Definition}



\subsection{Theorem \ref{theorem:local-validity-splitcp}}
\label{subsec:proof-local-validity-splicp}
\begin{proof}[Proof of Theorem \ref{theorem:local-validity-splitcp}]

Let $A_j \in \sA$ be an arbitrary partition element. Consider $I_j = \{ i \in I_{\cal} : \X_i \in A_j\}$. Since the full data $\{ (\X_i, Y_i) \}_{i \in I_\cal \sqcup \{n+1\}}$ is exchangeable, the scores $\hat{s}_i$ for $i \in I_j \sqcup \{n+1\}$ are exchangeable conditional on  $I_j$ and $\X_{n+1} \in A_j$. Therefore, by the definition of $s^*_j$,
\begin{align*}
    \P[Y_{n+1} \in C_{\local}(\X_{n+1}) | \X_{n+1} \in A_j, I_j ] &= \P[\hat{s}(\X_{n+1}, Y_{n+1}) \le s^*_j | \X_{n+1} \in A_j, I_j ] \\ 
    &\ge 1-\alpha.
\end{align*}
Because this holds for all $I_j$, the bound is true after marginalizing over $I_j$. Thus, 
\begin{equation}
\label{eq:local-guarantee}
    \P[Y_{n+1} \in C_{\local}(\X_{n+1}) | \X_{n+1} \in A_j ] \ge 1-\alpha.
\end{equation}
As this holds for an arbitrary element $A_j$, \eqref{eq:local-guarantee} is valid for all $j=1, \dots, K$. Corollary \ref{thm:local-Locart} is proved by applying Theorem \ref{theorem:local-validity-splitcp} in the \locart{} partition and Corollary \ref{thm:marginal-Locart} follows from \citeauthor{Lei2014}.\end{proof}

\subsection{Theorem \ref{thm:optimal_cutoff_eq_relation}}
\label{subsec:proof_optimal_partition}
\begin{proof}[Proof of Theorem \ref{thm:optimal_cutoff_eq_relation}]
    To prove item \ref{item_1_eq_relation} notice that if $\x_1, \x_2 \in A$, then $\hat{s}(\X, Y)|\X= \x_1 \sim \hat{s}(\X, Y)|\X = \x_2$, which by Definition \ref{def:oracle_PI} implies directly that $t_{1-\alpha}^*(\x_1) = t_{1-\alpha}^*(\x_2)$ for any $\alpha \in (0,1)$. To show item \ref{item_2_eq_relation}, assume that $\x_1, \x_2 \in J$, and then $t_{1-\alpha}^*(\x_1) = t_{1-\alpha}^*(\x_2)$ for every $\alpha \in (0,1)$. Conclude that $F_{\hat{s}(\X, Y)|\X= \x_1}^{-1}(1 - \alpha|\x_1) = F_{\hat{s}(\X, Y)|\X= \x_2}^{-1}(1 - \alpha|\x_2)$ for every $\alpha \in (0,1)$. Thus, from the monotonicity and continuity of $F_{{\hat{s}(\X, Y)|\X= \x}}$, we obtain that $F_{{\hat{s}(\X, Y)|\X= \x}_1}(r|\x_1) = F_{{\hat{s}(\X, Y)|\X= \x}_2}(r|\x_2)$ for all $r \in \R$. It follows that  $\hat{s}(\X, Y)|\X= \x_1 \sim \hat{s}(\X, Y)|\X= \x_2 $, and therefore $\x_1, \x_2 \in A$.
\end{proof}

\subsection{Theorem \ref{thm:consistency_locart}}
\label{subsec:proof_consistency_locart}
\begin{proof}[Proof of Theorem \ref{thm:consistency_locart}]
    Under Assumptions \ref{assump:CART_functions} and \ref{assump:continuity}, the theorem follows from Lemma 34 in \cite{izbicki2022cd}.
\end{proof}

\subsection{Theorem \ref{thm:asympt-cond-Locart}}
\label{subsec:proof_asympt_valid}
To prove Theorem \ref{thm:asympt-cond-Locart} we use the following Lemmas \ref{lemma:oracle_vs_PI} and \ref{lemma:oracle_vs_locart} proved below:
\begin{Lemma}
\label{lemma:oracle_vs_PI}
    If a prediction interval method $C(\cdot)$ converges to the oracle prediction interval $C^*(\cdot)$, then it satisfies asymptotic conditional coverage.
\end{Lemma}
\begin{proof}[Proof of Lemma \ref{lemma:oracle_vs_PI}]
    Since $\P[Y_{n + 1} \in C^*(\X_{n + 1}) \Delta C(\X_{n + 1})] = o(1)$, it follows from Markov's inequality and the dominated convergence theorem that 
    $$\P[Y_{n + 1} \in C^*(\X_{n + 1}) \Delta C(\X_{n + 1})|\X_{n + 1}] = o_{\P}(1) \; .$$ 
    Therefore, there exists $\phi_n = o(1)$ such that, for $$\Lambda_{n + 1}^{c} = \{ \x_{n + 1} \in \sX: \P[Y_{n + 1} \in C^*(\X_{n + 1}) \Delta C(\X_{n + 1})|\X_{n + 1} = \x_{n + 1}] > \phi_n\} \; ,$$
    one obtains $\P[\X_{n + 1} \in \Lambda_{n + 1}^c] = o(1)$. Conclude that $\P[\X_{n + 1} \in \Lambda_{n + 1}] = 1 - o(1)$ and that:
    \begin{align*}
        &\sup_{\x_{n + 1} \in \Lambda_n} \Big| \P[Y_{n + 1} \in C(\X_{n + 1})| \X_{n + 1} = \x_{n + 1}] - (1 - \alpha) \Big| \\
        &\leq \sup_{\x_{n + 1} \in \Lambda_n}  \Big| \P[Y_{n + 1} \in C^*(\X_{n + 1}) \Delta C(\X_{n + 1})|\X_{n + 1} = \x_{n + 1}] \Big| \leq \phi_n = o(1) .
    \end{align*}
\end{proof}
\begin{Lemma}
\label{lemma:oracle_vs_locart}
    Let $C^* = (\hat{\mu}(\X_{n + 1}) - t^{*}_{1-\alpha}(\X_{n + 1}), \hat{\mu}(\X_{n + 1}) + t^{*}_{1-\alpha}(\X_{n + 1}))$ and $C_{\locart{}} = (\hat{\mu}(\X_{n + 1}) - \hat{t}_{1-\alpha}(\X_{n + 1}), \hat{\mu}(\X_{n + 1}) + \hat{t}_{1-\alpha}(\X_{n + 1}))$, where $\hat{t}(.)$ is the \locart{} estimated cutoff for a fixed induced partition. Under Assumption \ref{assump:CART_functions}:
    \begin{align*}
        \P[Y_{n + 1} \in C^* \Delta C_{\locart{}}| \X_{n + 1}] = o_{\P}(1) \; .
    \end{align*}
\end{Lemma}
\begin{proof}[Proof of Lemma \ref{lemma:oracle_vs_locart}]
    Under Assumption \ref{assump:CART_functions}, $|t_{1-\alpha}^*(\X_{n + 1}) - \hat{t}_{1-\alpha}(\X_{n + 1})| = o_{\P}(1)$ by Theorem \ref{thm:consistency_locart}. Thus, there exists $\lambda_n = o(1)$ such that $\P[|t_{1-\alpha}^*(\X_{n + 1}) - \hat{t}_{1-\alpha}(\X_{n + 1})| > \lambda_n] = o(1)$. Note that, $\{Y_{n + 1} \in C^* \Delta C_{\locart{}} \} \subseteq \{|t_{1-\alpha}^*(\X_{n + 1}) - \hat{t}_{1-\alpha}(\X_{n + 1})| > \lambda_n\}$. Thus:
    \begin{align*} 
    \P[Y_{n + 1} \in C^* \Delta C_{\locart{}}] &\leq \P[|t_{1-\alpha}^*(\X_{n + 1}) - \hat{t}_{1-\alpha}(\X_{n + 1})| > \lambda_n] \\
    &= o(1) \, .
    \end{align*}
    Since $\P[Y_{n + 1} \in C^* \Delta C_{\locart{}}] = o(1)$ if follows from Markov's inequality that $\P[Y_{n + 1} \in C^* \Delta C_{\locart{}}| \X_{n + 1}] = o_{\P}(1)$.
\end{proof}
Now, we prove Theorem \ref{thm:asympt-cond-Locart}.
\begin{proof}[Proof of Theorem \ref{thm:asympt-cond-Locart}]
    Follows directly from Lemmas \ref{lemma:oracle_vs_PI} and \ref{lemma:oracle_vs_locart}.
\end{proof}

\section{Experiments additional figures and tables}
\label{sec:appendix_a}
\begin{figure}[H]
    \centering
    \includegraphics[width=\textwidth]{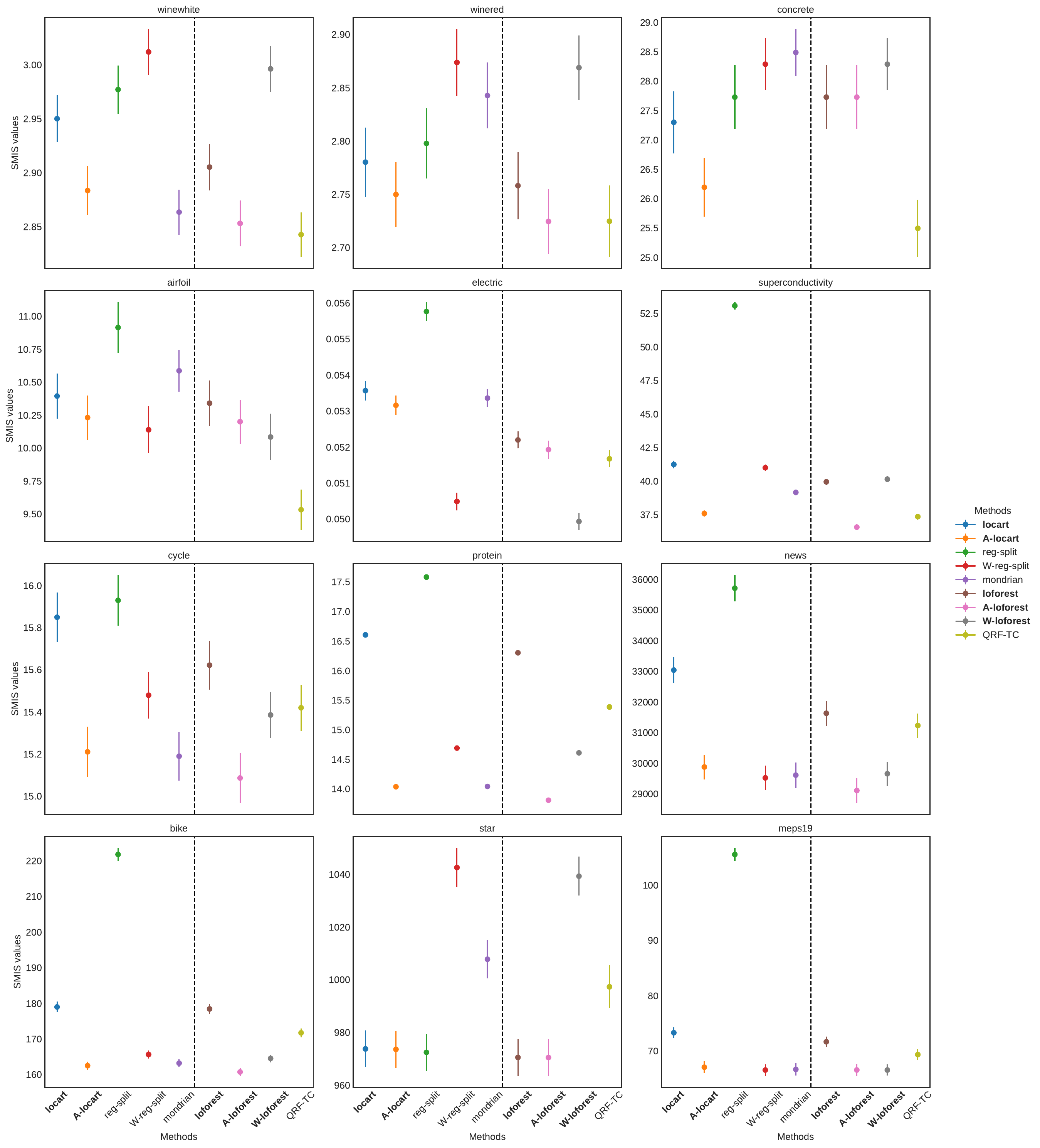}
    \caption{Smis 95\% confidence interval of each method for each real dataset. The black dotted line separates the conformal from the non-conformal methods. We observe that \alocart{} and \aloforest{} stand out in several datasets and have good performances in general.}
    \label{fig:smis_performance}
\end{figure}

\begin{figure}[H]
    \centering
    \includegraphics[width=\textwidth]{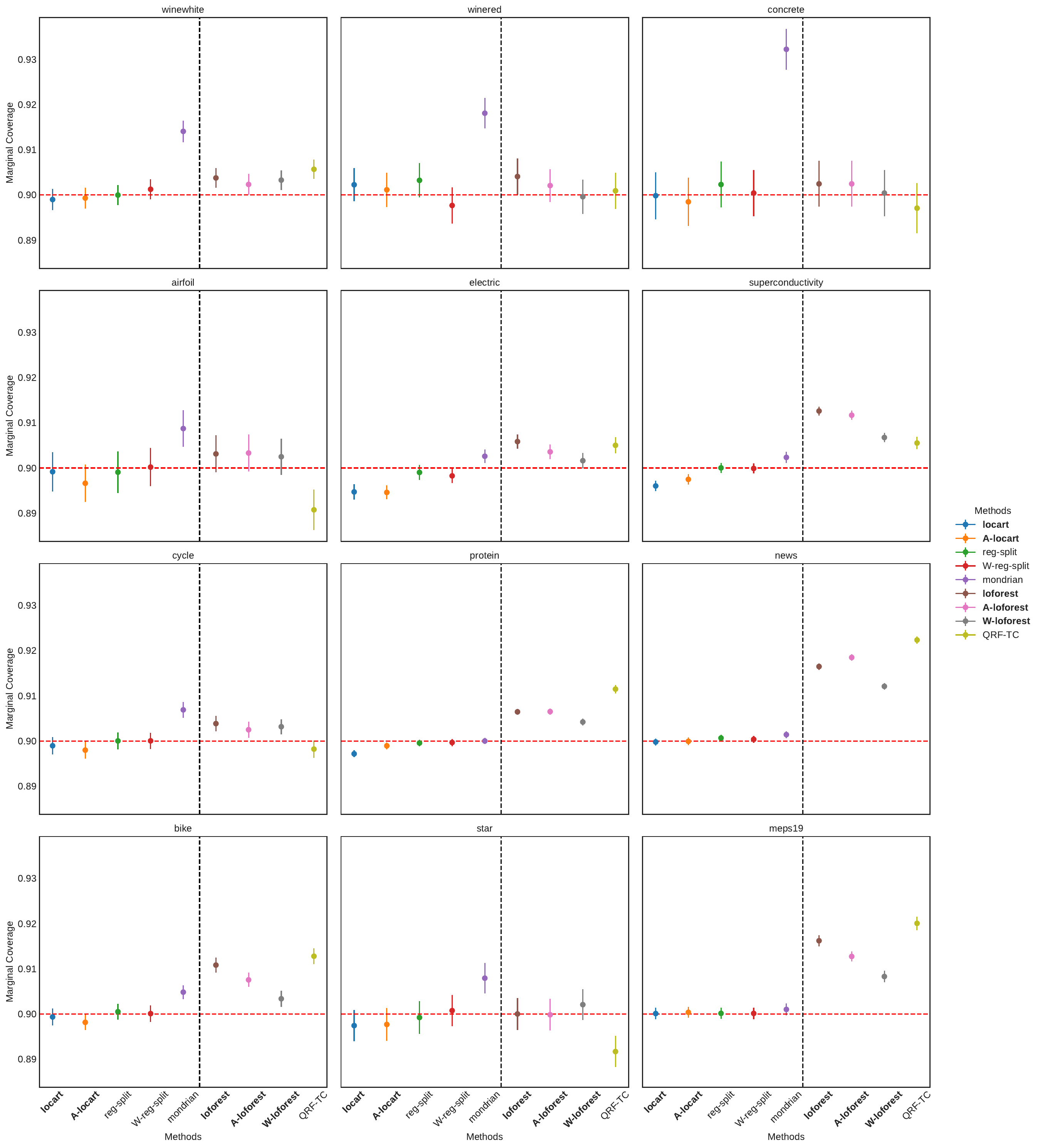}
    \caption{Marginal coverage 95\% confidence interval of each method for each real dataset. The red dotted line indicates the nominal level ($1 - \alpha = 0.9$) and the black dotted line separates the conformal from the non-conformal methods. In general, all conformal methods are close to the nominal level for each real dataset, while the non-conformal ones present slight over-coverage in some of the datasets.}
    \label{fig:marginal_coverage_real}
\end{figure}

\begin{table}[!hbt]
\centering
\caption{Mean marginal coverage values of each method for each kind of simulation. The average across 100 simulations is reported with two times the standard error in brackets. Values in bold indicate methods with the most distant marginal coverage from the nominal ($1 - \alpha = 0.9$).}
\label{tab:marginal_coverage_sim}
\adjustbox{width = 1.0\textwidth}{
\begin{tabular}{ccccccc|cccc}
\hline
kind                           & p & \locart{}                                        & \alocart{}                                        & RegSplit                                                & WRegSplit                                              & Mondrian                                                 & \loforest{}                                                 & \aloforest{}                                               & \wloforest{}                                               & QRF-TC                                                            \\ \hline
Asym.                     & 1 & \begin{tabular}[c]{@{}c@{}}0.897 \\ (0.001)\end{tabular} & \begin{tabular}[c]{@{}c@{}}0.898 \\ (0.001)\end{tabular} & \begin{tabular}[c]{@{}c@{}}0.898 \\ (0.001)\end{tabular} & \begin{tabular}[c]{@{}c@{}}0.898 \\ (0.001)\end{tabular} & \begin{tabular}[c]{@{}c@{}}0.9 \\ (0.001)\end{tabular}   & \begin{tabular}[c]{@{}c@{}}0.903 \\ (0.001)\end{tabular}          & \textbf{\begin{tabular}[c]{@{}c@{}}0.906 \\ (0.001)\end{tabular}} & \begin{tabular}[c]{@{}c@{}}0.905 \\ (0.001)\end{tabular}          & \textbf{\begin{tabular}[c]{@{}c@{}}0.906 \\ (0.001)\end{tabular}} \\
Asym.                     & 3 & \begin{tabular}[c]{@{}c@{}}0.897 \\ (0.001)\end{tabular} & \begin{tabular}[c]{@{}c@{}}0.898 \\ (0.001)\end{tabular} & \begin{tabular}[c]{@{}c@{}}0.898 \\ (0.001)\end{tabular} & \begin{tabular}[c]{@{}c@{}}0.898 \\ (0.001)\end{tabular} & \begin{tabular}[c]{@{}c@{}}0.9 \\ (0.001)\end{tabular}   & \textbf{\begin{tabular}[c]{@{}c@{}}0.906 \\ (0.001)\end{tabular}} & \begin{tabular}[c]{@{}c@{}}0.905 \\ (0.001)\end{tabular}          & \begin{tabular}[c]{@{}c@{}}0.906 \\ (0.001)\end{tabular}          & \begin{tabular}[c]{@{}c@{}}0.9 \\ (0.002)\end{tabular}            \\
Asym.                     & 5 & \begin{tabular}[c]{@{}c@{}}0.898 \\ (0.001)\end{tabular} & \begin{tabular}[c]{@{}c@{}}0.899 \\ (0.001)\end{tabular} & \begin{tabular}[c]{@{}c@{}}0.899 \\ (0.001)\end{tabular} & \begin{tabular}[c]{@{}c@{}}0.898 \\ (0.001)\end{tabular} & \begin{tabular}[c]{@{}c@{}}0.9 \\ (0.001)\end{tabular}   & \begin{tabular}[c]{@{}c@{}}0.905 \\ (0.001)\end{tabular}          & \begin{tabular}[c]{@{}c@{}}0.905 \\ (0.001)\end{tabular}          & \textbf{\begin{tabular}[c]{@{}c@{}}0.906 \\ (0.001)\end{tabular}} & \begin{tabular}[c]{@{}c@{}}0.898 \\ (0.002)\end{tabular}          \\ \hline
Asym. 2                   & 1 & \begin{tabular}[c]{@{}c@{}}0.897 \\ (0.001)\end{tabular} & \begin{tabular}[c]{@{}c@{}}0.898 \\ (0.001)\end{tabular} & \begin{tabular}[c]{@{}c@{}}0.9 \\ (0.001)\end{tabular}   & \begin{tabular}[c]{@{}c@{}}0.899 \\ (0.001)\end{tabular} & \begin{tabular}[c]{@{}c@{}}0.9 \\ (0.001)\end{tabular}   & \begin{tabular}[c]{@{}c@{}}0.904 \\ (0.001)\end{tabular}          & \begin{tabular}[c]{@{}c@{}}0.906 \\ (0.001)\end{tabular}          & \begin{tabular}[c]{@{}c@{}}0.905 \\ (0.001)\end{tabular}          & \textbf{\begin{tabular}[c]{@{}c@{}}0.913 \\ (0.001)\end{tabular}} \\
Asym. 2                   & 3 & \begin{tabular}[c]{@{}c@{}}0.893 \\ (0.001)\end{tabular} & \begin{tabular}[c]{@{}c@{}}0.897 \\ (0.001)\end{tabular} & \begin{tabular}[c]{@{}c@{}}0.896 \\ (0.001)\end{tabular} & \begin{tabular}[c]{@{}c@{}}0.897 \\ (0.001)\end{tabular} & \begin{tabular}[c]{@{}c@{}}0.898 \\ (0.001)\end{tabular} & \textbf{\begin{tabular}[c]{@{}c@{}}0.911 \\ (0.001)\end{tabular}} & \begin{tabular}[c]{@{}c@{}}0.904 \\ (0.001)\end{tabular}          & \begin{tabular}[c]{@{}c@{}}0.904 \\ (0.001)\end{tabular}          & \begin{tabular}[c]{@{}c@{}}0.906 \\ (0.001)\end{tabular}          \\
Asym. 2                   & 5 & \begin{tabular}[c]{@{}c@{}}0.899 \\ (0.001)\end{tabular} & \begin{tabular}[c]{@{}c@{}}0.899 \\ (0.001)\end{tabular} & \begin{tabular}[c]{@{}c@{}}0.9 \\ (0.001)\end{tabular}   & \begin{tabular}[c]{@{}c@{}}0.9 \\ (0.001)\end{tabular}   & \begin{tabular}[c]{@{}c@{}}0.901 \\ (0.001)\end{tabular} & \textbf{\begin{tabular}[c]{@{}c@{}}0.911 \\ (0.001)\end{tabular}} & \begin{tabular}[c]{@{}c@{}}0.908 \\ (0.001)\end{tabular}          & \begin{tabular}[c]{@{}c@{}}0.908 \\ (0.001)\end{tabular}          & \begin{tabular}[c]{@{}c@{}}0.904 \\ (0.001)\end{tabular}          \\ \hline
Heterosc.                & 1 & \begin{tabular}[c]{@{}c@{}}0.898 \\ (0.001)\end{tabular} & \begin{tabular}[c]{@{}c@{}}0.899 \\ (0.001)\end{tabular} & \begin{tabular}[c]{@{}c@{}}0.898 \\ (0.001)\end{tabular} & \begin{tabular}[c]{@{}c@{}}0.901 \\ (0.001)\end{tabular} & \begin{tabular}[c]{@{}c@{}}0.902 \\ (0.001)\end{tabular} & \begin{tabular}[c]{@{}c@{}}0.902 \\ (0.001)\end{tabular}          & \textbf{\begin{tabular}[c]{@{}c@{}}0.903 \\ (0.001)\end{tabular}} & \begin{tabular}[c]{@{}c@{}}0.902 \\ (0.001)\end{tabular}          & \begin{tabular}[c]{@{}c@{}}0.902 \\ (0.001)\end{tabular}          \\
Heterosc.                & 3 & \begin{tabular}[c]{@{}c@{}}0.896 \\ (0.001)\end{tabular} & \begin{tabular}[c]{@{}c@{}}0.9 \\ (0.001)\end{tabular}   & \begin{tabular}[c]{@{}c@{}}0.899 \\ (0.001)\end{tabular} & \begin{tabular}[c]{@{}c@{}}0.9 \\ (0.001)\end{tabular}   & \begin{tabular}[c]{@{}c@{}}0.902 \\ (0.001)\end{tabular} & \textbf{\begin{tabular}[c]{@{}c@{}}0.906 \\ (0.001)\end{tabular}} & \begin{tabular}[c]{@{}c@{}}0.902 \\ (0.001)\end{tabular}          & \begin{tabular}[c]{@{}c@{}}0.902 \\ (0.001)\end{tabular}          & \begin{tabular}[c]{@{}c@{}}0.9 \\ (0.001)\end{tabular}            \\
Heterosc.                & 5 & \begin{tabular}[c]{@{}c@{}}0.899 \\ (0.001)\end{tabular} & \begin{tabular}[c]{@{}c@{}}0.9 \\ (0.001)\end{tabular}   & \begin{tabular}[c]{@{}c@{}}0.9 \\ (0.001)\end{tabular}   & \begin{tabular}[c]{@{}c@{}}0.9 \\ (0.001)\end{tabular}   & \begin{tabular}[c]{@{}c@{}}0.902 \\ (0.001)\end{tabular} & \textbf{\begin{tabular}[c]{@{}c@{}}0.904 \\ (0.001)\end{tabular}} & \begin{tabular}[c]{@{}c@{}}0.902 \\ (0.001)\end{tabular}          & \begin{tabular}[c]{@{}c@{}}0.902 \\ (0.001)\end{tabular}          & \begin{tabular}[c]{@{}c@{}}0.899 \\ (0.002)\end{tabular}          \\ \hline
Homosc.                 & 1 & \begin{tabular}[c]{@{}c@{}}0.9 \\ (0.001)\end{tabular}   & \begin{tabular}[c]{@{}c@{}}0.9 \\ (0.001)\end{tabular}   & \begin{tabular}[c]{@{}c@{}}0.9 \\ (0.001)\end{tabular}   & \begin{tabular}[c]{@{}c@{}}0.901 \\ (0.001)\end{tabular} & \begin{tabular}[c]{@{}c@{}}0.902 \\ (0.001)\end{tabular} & \begin{tabular}[c]{@{}c@{}}0.901 \\ (0.001)\end{tabular}          & \begin{tabular}[c]{@{}c@{}}0.901 \\ (0.001)\end{tabular}          & \begin{tabular}[c]{@{}c@{}}0.902 \\ (0.001)\end{tabular}          & \textbf{\begin{tabular}[c]{@{}c@{}}0.897 \\ (0.002)\end{tabular}} \\
Homosc.                 & 3 & \begin{tabular}[c]{@{}c@{}}0.901 \\ (0.001)\end{tabular} & \begin{tabular}[c]{@{}c@{}}0.901 \\ (0.001)\end{tabular} & \begin{tabular}[c]{@{}c@{}}0.901 \\ (0.001)\end{tabular} & \begin{tabular}[c]{@{}c@{}}0.901 \\ (0.001)\end{tabular} & \begin{tabular}[c]{@{}c@{}}0.903 \\ (0.001)\end{tabular} & \begin{tabular}[c]{@{}c@{}}0.902 \\ (0.001)\end{tabular}          & \begin{tabular}[c]{@{}c@{}}0.902 \\ (0.001)\end{tabular}          & \begin{tabular}[c]{@{}c@{}}0.902 \\ (0.001)\end{tabular}          & \textbf{\begin{tabular}[c]{@{}c@{}}0.896 \\ (0.002)\end{tabular}} \\
Homosc.                 & 5 & \begin{tabular}[c]{@{}c@{}}0.901 \\ (0.001)\end{tabular} & \begin{tabular}[c]{@{}c@{}}0.901 \\ (0.001)\end{tabular} & \begin{tabular}[c]{@{}c@{}}0.901 \\ (0.001)\end{tabular} & \begin{tabular}[c]{@{}c@{}}0.901 \\ (0.001)\end{tabular} & \begin{tabular}[c]{@{}c@{}}0.902 \\ (0.001)\end{tabular} & \begin{tabular}[c]{@{}c@{}}0.902 \\ (0.001)\end{tabular}          & \begin{tabular}[c]{@{}c@{}}0.901 \\ (0.001)\end{tabular}          & \begin{tabular}[c]{@{}c@{}}0.903 \\ (0.001)\end{tabular}          & \textbf{\begin{tabular}[c]{@{}c@{}}0.895 \\ (0.001)\end{tabular}} \\ \hline
Non-corr. heterosc. & 1 & \begin{tabular}[c]{@{}c@{}}0.898 \\ (0.001)\end{tabular} & \begin{tabular}[c]{@{}c@{}}0.898 \\ (0.001)\end{tabular} & \begin{tabular}[c]{@{}c@{}}0.898 \\ (0.001)\end{tabular} & \begin{tabular}[c]{@{}c@{}}0.9 \\ (0.001)\end{tabular}   & \begin{tabular}[c]{@{}c@{}}0.9 \\ (0.001)\end{tabular}   & \textbf{\begin{tabular}[c]{@{}c@{}}0.902 \\ (0.001)\end{tabular}} & \textbf{\begin{tabular}[c]{@{}c@{}}0.902 \\ (0.001)\end{tabular}} & \textbf{\begin{tabular}[c]{@{}c@{}}0.902 \\ (0.001)\end{tabular}} & \textbf{\begin{tabular}[c]{@{}c@{}}0.902 \\ (0.001)\end{tabular}} \\
Non-corr. heterosc. & 3 & \begin{tabular}[c]{@{}c@{}}0.896 \\ (0.001)\end{tabular} & \begin{tabular}[c]{@{}c@{}}0.897 \\ (0.001)\end{tabular} & \begin{tabular}[c]{@{}c@{}}0.899 \\ (0.001)\end{tabular} & \begin{tabular}[c]{@{}c@{}}0.9 \\ (0.001)\end{tabular}   & \begin{tabular}[c]{@{}c@{}}0.901 \\ (0.001)\end{tabular} & \textbf{\begin{tabular}[c]{@{}c@{}}0.906 \\ (0.001)\end{tabular}} & \begin{tabular}[c]{@{}c@{}}0.904 \\ (0.001)\end{tabular}          & \begin{tabular}[c]{@{}c@{}}0.902 \\ (0.001)\end{tabular}          & \begin{tabular}[c]{@{}c@{}}0.9 \\ (0.001)\end{tabular}            \\
Non-corr. heterosc. & 5 & \begin{tabular}[c]{@{}c@{}}0.898 \\ (0.001)\end{tabular} & \begin{tabular}[c]{@{}c@{}}0.899 \\ (0.001)\end{tabular} & \begin{tabular}[c]{@{}c@{}}0.9 \\ (0.001)\end{tabular}   & \begin{tabular}[c]{@{}c@{}}0.9 \\ (0.001)\end{tabular}   & \begin{tabular}[c]{@{}c@{}}0.901 \\ (0.001)\end{tabular} & \textbf{\begin{tabular}[c]{@{}c@{}}0.904 \\ (0.001)\end{tabular}} & \begin{tabular}[c]{@{}c@{}}0.902 \\ (0.001)\end{tabular}          & \begin{tabular}[c]{@{}c@{}}0.902 \\ (0.001)\end{tabular}          & \begin{tabular}[c]{@{}c@{}}0.898 \\ (0.002)\end{tabular}          \\ \hline
$t$-residuals                    & 1 & \begin{tabular}[c]{@{}c@{}}0.9 \\ (0.001)\end{tabular}   & \begin{tabular}[c]{@{}c@{}}0.9 \\ (0.001)\end{tabular}   & \begin{tabular}[c]{@{}c@{}}0.9 \\ (0.001)\end{tabular}   & \begin{tabular}[c]{@{}c@{}}0.901 \\ (0.001)\end{tabular} & \begin{tabular}[c]{@{}c@{}}0.902 \\ (0.001)\end{tabular} & \begin{tabular}[c]{@{}c@{}}0.903 \\ (0.001)\end{tabular}          & \textbf{\begin{tabular}[c]{@{}c@{}}0.904 \\ (0.001)\end{tabular}} & \textbf{\begin{tabular}[c]{@{}c@{}}0.904 \\ (0.001)\end{tabular}} & \begin{tabular}[c]{@{}c@{}}0.897 \\ (0.002)\end{tabular}          \\
$t$-residuals                    & 3 & \begin{tabular}[c]{@{}c@{}}0.901 \\ (0.001)\end{tabular} & \begin{tabular}[c]{@{}c@{}}0.901 \\ (0.001)\end{tabular} & \begin{tabular}[c]{@{}c@{}}0.901 \\ (0.001)\end{tabular} & \begin{tabular}[c]{@{}c@{}}0.901 \\ (0.001)\end{tabular} & \begin{tabular}[c]{@{}c@{}}0.903 \\ (0.001)\end{tabular} & \begin{tabular}[c]{@{}c@{}}0.904 \\ (0.001)\end{tabular}          & \begin{tabular}[c]{@{}c@{}}0.904 \\ (0.001)\end{tabular}          & \textbf{\begin{tabular}[c]{@{}c@{}}0.905 \\ (0.001)\end{tabular}} & \begin{tabular}[c]{@{}c@{}}0.898 \\ (0.002)\end{tabular}          \\
$t$-residuals                    & 5 & \begin{tabular}[c]{@{}c@{}}0.901 \\ (0.001)\end{tabular} & \begin{tabular}[c]{@{}c@{}}0.901 \\ (0.001)\end{tabular} & \begin{tabular}[c]{@{}c@{}}0.901 \\ (0.001)\end{tabular} & \begin{tabular}[c]{@{}c@{}}0.902 \\ (0.001)\end{tabular} & \begin{tabular}[c]{@{}c@{}}0.903 \\ (0.001)\end{tabular} & \begin{tabular}[c]{@{}c@{}}0.904 \\ (0.001)\end{tabular}          & \begin{tabular}[c]{@{}c@{}}0.904 \\ (0.001)\end{tabular}          & \textbf{\begin{tabular}[c]{@{}c@{}}0.906 \\ (0.001)\end{tabular}} & \begin{tabular}[c]{@{}c@{}}0.898 \\ (0.002)\end{tabular}          \\ \hline
\end{tabular}
}
\end{table}





\bibliographystyle{elsarticle-num-names}
\bibliography{main}
 
\end{document}